\documentclass{article}

\usepackage[preprint]{neurips_2022}

\usepackage{caption}
\usepackage{subcaption}

\usepackage{wrapfig}
\usepackage[final]{graphicx}

\usepackage[utf8]{inputenc} 
\usepackage[T1]{fontenc}    
\usepackage{hyperref}       
\usepackage{url}            
\usepackage{booktabs}       
\usepackage{amsfonts}       
\usepackage{nicefrac}       
\usepackage{microtype}      
\usepackage{xcolor}         
\usepackage{bm,bbm}
\usepackage{amsmath}
\usepackage{graphicx}
\usepackage{algorithm}
\usepackage{algorithmic}
\usepackage{listings}
\usepackage{multirow}
\usepackage{wrapfig}

\usepackage[english]{babel}
\usepackage{amsthm}

\def \R {\mathbb{R}}

\newtheorem{theorem}{Theorem}

\title{TreeDRNet: A Robust Deep Model for Long Term Time Series Forecasting}

\author{Tian Zhou$^*$\And Jianqing zhu$^*$\And Xue Wang\And Ziqing Ma\And Qingsong Wen\And Liang Sun\And Rong Jin}

\begin{document}

\maketitle

\begin{abstract}
Various deep learning models, especially some latest Transformer-based approaches, have greatly improved the state-of-art performance for long-term time series forecasting. However, those transformer-based models suffer a severe deterioration performance with prolonged input length, which prohibits them from using extended historical info. Moreover, these methods tend to handle complex examples in long-term forecasting with increased model complexity, which often leads to a significant increase in computation and less robustness in performance (e.g., overfitting). We propose a novel neural network architecture, called TreeDRNet, for more effective long-term forecasting. Inspired by robust regression, we introduce doubly residual link structure to make prediction more robust. Built upon Kolmogorov–Arnold representation theorem, we explicitly introduce feature selection, model ensemble, and a tree structure to further utilize the extended input sequence, which improves the robustness and representation power of TreeDRNet. Unlike previous deep models for sequential forecasting work, TreeDRNet is built entirely on multilayer perceptron and thus enjoys high computational efficiency. Our extensive empirical studies show that TreeDRNet is significantly more effective than state-of-the-art methods, reducing prediction errors by $20\%$ to $40\%$ for multivariate time series. In particular, TreeDRNet is over $10$ times more efficient than transformer-based methods. The code will be released soon.\footnote{$^*$ Equal contribution}


\end{abstract}

\section{Introduction}

Long-term time series forecasting has played an important role in numerous applications, such as retail~\cite{bose2017probabilistic,courty1999timing}
, healthcare~\cite{lim2018forecasting,zhang2018multi}, and engineering systems~\cite{zhang2019deep,gonzalez2019methodology}. Various deep models have been developed for sequential forecasting, and among which recurrent neural network is probably mostly well examined~\cite{connor1994recurrent,hewamalage2021recurrent}. Following the recent success in natural language process (NLP) and computer vision (CV) communities~\cite{attention_is_all_you_need,Bert/NAACL/Jacob,Transformers-for-image-at-scale/iclr/DosovitskiyB0WZ21,DBLP:Global-filter-FNO-in-cv}, Transformer~\cite{attention_is_all_you_need} has been introduced to capture long-term dependencies in time series forecasting and shows promising results~\cite{haoyietal-informer-2021,Autoformer}.

\begin{figure}[t]
\centering
\begin{minipage}[t]{0.48\textwidth}
\centering
\includegraphics[width=\textwidth]{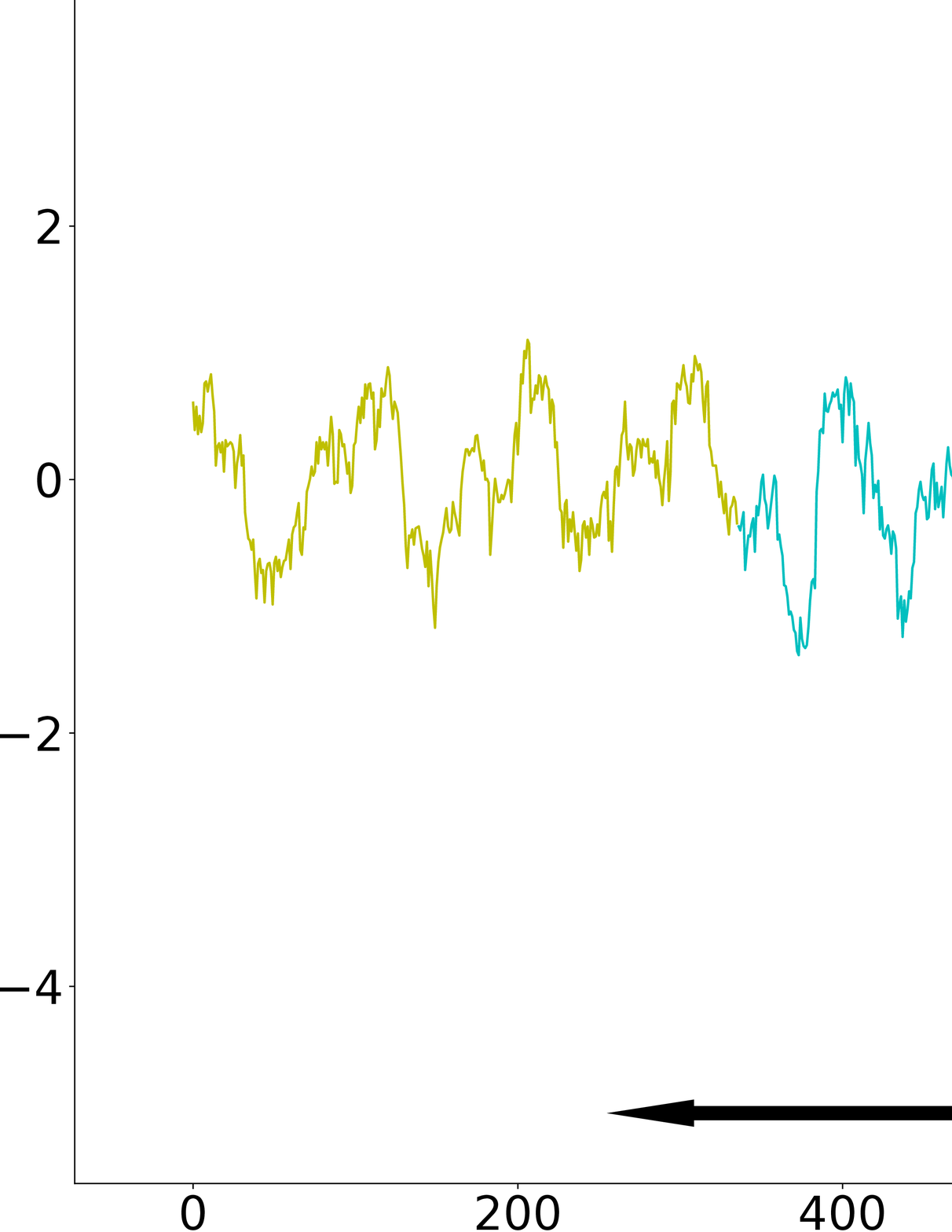}
\end{minipage}
\begin{minipage}[t]{0.48\textwidth}
\centering
\includegraphics[width=\textwidth]{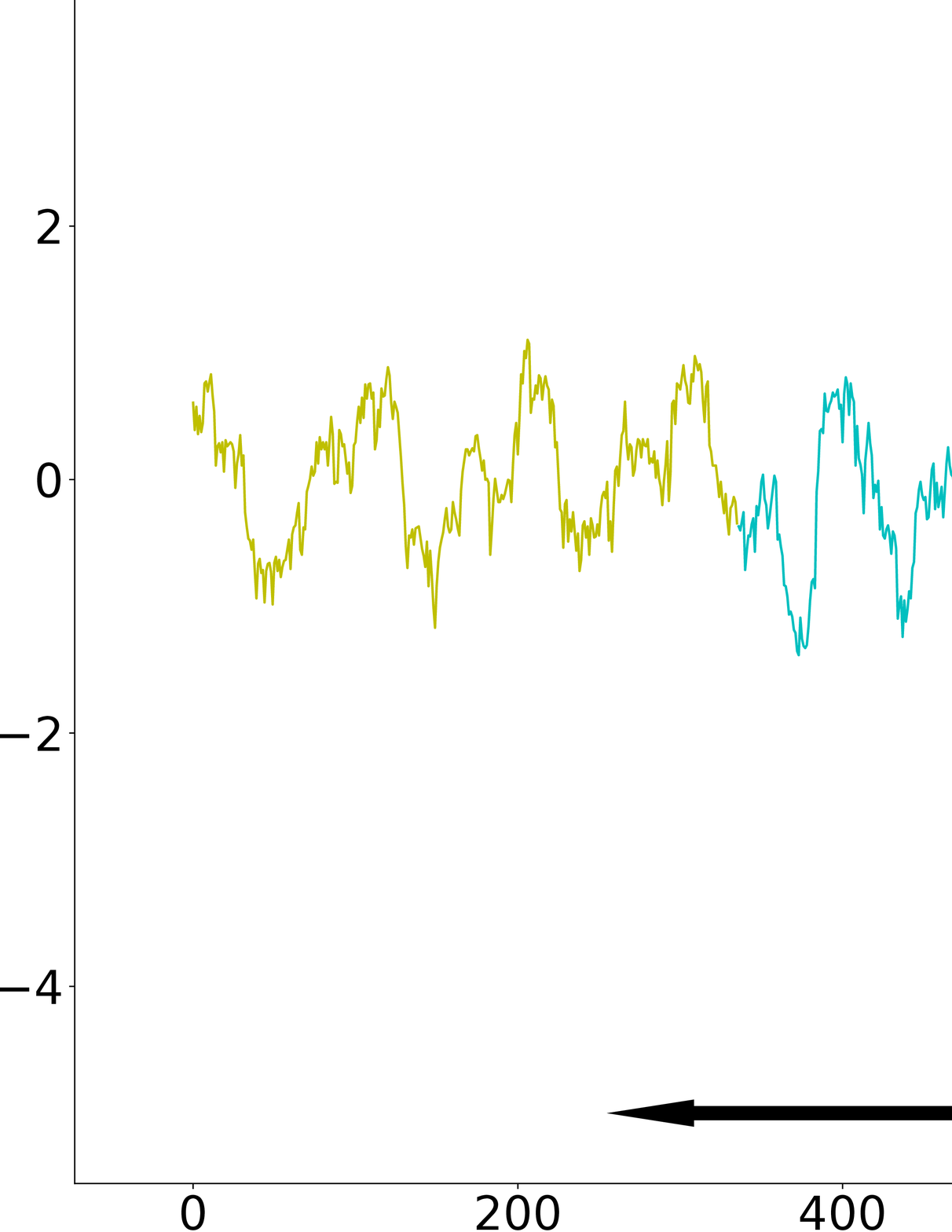}
\end{minipage}
\caption{Left: TreeDRNet forecasting sequences with model trained using  different input horizons. Right: Autoformer forecasting sequences with model trained using different input horizons. Input horizons: the yellow (also including the blue and purple) is 1008; the blue (also including the purple) is 672; the purple is 336. }
\label{Robutness}
\end{figure}

One of the main challenges faced by long-term sequential prediction is robustness. There are two kinds of robustness here: one is robust to a long noisy input, which facilitates the model to learn more information from a noisy extended historical sequence. The other is robust forecasting of various long outputs using a short fixed input sequence. Most recent Transformer-based models fail on the former case, which makes models' output distribution change significantly and performance drop tremendously with increasing input length as shown in Figure \ref{Robutness}. They cannot use extended historical data for better forecasting. Moreover, complicated models, such as Transformer, are introduced to cover complex examples in long-term forecasting, which leads to a significant increase in computation and potential overfitting of training data, as we observed in our studies. 

In this work, we propose a robust deep model for long-term forecasting, termed {\bf TreeDRNet}. Our model is inspired by the iteratively reweighted algorithm for robust regression~\cite{daubechies2010iteratively}, in which we introduce a doubly residual link structure to improve the model robustness. To further improve the reliability and accuracy of long-term forecasting, we borrow insights from Kolmogorov–Arnold representation theorem~\cite{hecht1987kolmogorov}, based on which feature selection, model ensemble, and a tree structure are integrated into TreeDRNet. As we demonstrated in our empirical studies, it is the combination of the four key features (i.e., doubly residual links, feature selection, model ensemble, and tree structure) that makes TreeDRNet more effective than state-of-the-art methods for long-term forecasting, reducing the prediction error by $20\%$ to $40\%$ for multivariate forecasting. Besides its prediction power, the entire TreeDRNet is built upon multilayer perceptron (MLP) and thus enjoys high computational efficiency. For instance, compared to state-of-the-art (SOTA) Transformer-based forecasting models, TreeDRNet 
exhibits over $10$ times more efficiency for both model training and inference.

In short, the contributions of this work can be mainly summarized as follows.
\begin{itemize}

\item Inspired by the iteratively reweighted algorithm in robust regression and Kolmogorov–Arnold representation theorem, we propose a robust long-term prediction deep model TreeDRNet, which integrates doubly residual links, feature selection, model ensemble, and tree structure to achieve the state-of-the-art results for long-term forecasting. 

\item The proposed TreeDRNet is built entirely based on MLP and thus enjoys high computational efficiency compared with complicated forecasting models, with over $10$ times more efficient than Transformer-based methods.

\item We conduct extensive experiments on multiple benchmark datasets across diverse domains and show that the proposed TreeDRNet model improves the performance of the state-of-the-art methods significantly by reducing prediction errors by 20\% to 40\% for long-term multivariate time series forecasting. 

   
\end{itemize}


\section{Related work}
The common use of deep networks in forecasting is based on the RNN structure. The Multi-horizon Quantile Recurrent Forecaster (MQRNN) \cite{wen2017multi} uses LSTM building block to generate context vectors and then feed into MLPs to produce forecast value for each step. DeepAR \cite{salinas2020deepar}  uses the stacking LSTM layers to generate a Gaussian predictive distribution. \textcolor{black}{\cite{guo2019exploring}} \cite{fan2019multi} use the LSTM-based encoder with an attention mechanism to construct context vectors for a bidirectional LSTM-based decoder. However, LSTM belongs to a recursive time series forecasting method. The previous step predictions would be used as inputs for future predictions, so the prediction errors in the network will accumulate gradually. In contrast, our TreeDRNet adopts fully connected layers to process all time steps information and forecast the full horizon targets in one forward path. 
Recently, N-BEATS~\cite{oreshkin2019n} is proposed by designing a simple doubly residual sequential neural network model with multilayer perceptrons, which is also computationally efficient and achieves good results on univariate time series forecasting. 
One difference between our TreeDRNet and N-BEATS is that our model is theoretically motivated by the iterative reweighted algorithm in sparse recovery to design the doubly residual network. Another difference is that TreeDRNet borrows insights from Kolmogorov–Arnold representation theorem and constructs a novel tree-based neural network where each node is further enhanced with feature selection and model ensemble to aggregate the information effectively.


With the innovation of Transformers in NLP and CV~\cite{attention_is_all_you_need,Bert/NAACL/Jacob,Transformers-for-image-at-scale/iclr/DosovitskiyB0WZ21}, Transformer-based models are also proposed and achieved state-of-the-art performance in time series forecasting~\cite{Log-transformer-shiyang-2019,haoyietal-informer-2021,Autoformer}, since Transformer-based models excel in modeling long-term dependencies for sequential data. 
Furthermore, many efficient Transformers are designed in time series forecasting to overcome the quadratic computation complexity of the original Transformer without performance degradation. LogTrans~\cite{Log-transformer-shiyang-2019} adopts log-sparse attention and achieves $N\log^2 N$ complexity. Reformer~\cite{DBLP:conf/iclr/KitaevKL20-reformer} introduces a local-sensitive hashing that reduces the complexity to $N\log N$. Informer~\cite{haoyietal-informer-2021} uses a KL-divergence-based method to select top-k in the attention matrix and costs $N\log N$ in complexity. Most recently, Autoformer~\cite{Autoformer} introduced an auto-correlation block in place of canonical attention to perform the sub-series level attention, which achieves $N\log N$ complexity with the help of Fast Fourier transform (FFT) and top-k selection in the auto-correlation matrix.

Despite the progress made in Efficient-Transformers for time series modeling, these models are still with high computational complexity. They cannot appropriately deal with complex cases in long-term forecasting. In contrast, our TreeDRNet model is built entirely based on MLP with much less computational complexity while achieving impressive performance than transformer-based models in long-term forecasting scenarios.




\label{App:related_work_transformers}

\section{Motivation and Theoretical Foundation}\label{sec_motivation_theory}

In this section, we introduce the theory which motivated the model design. First, the double residual link structure can be interpreted as a generalization of the iterative reweighted algorithm, which exhibits good robustness. Next, we introduce the Kolmogorlv-Arnold representation to show that the two ensemble techniques (Model Ensemble and Tree-based Aggregation) employed in our framework enforce the model's expressive power.

\subsection{Double Residual Link Structure: A Generalization of Iterative Reweighted Algorithm}

We motivate the design of our double residual link structure by first stating the iterative reweighted algorithm~\cite{daubechies2010iteratively} (details in Appendix \ref{app:iter_rewei}) which is widely used in sparse recovery. Consider a simple linear regression problem. Let $(x_i, y_i), i=1, \ldots, n$ be the set of training examples, where $x_i \in \R^d$ and $y_i \in \R$. Our goal is to find a linear regression solution $\beta \in \R^d$ that is robust to error in $y$. A common approach toward robust regression problems is introducing $\ell_p$ as the loss function, where $p \in (0, 1]$. It amounts to solving the following non-convex optimization problem.
\[
    \min\limits_{\beta} \mathcal{L}(\beta) = \sum_{i=1}^n \left|\langle \beta, x_i \rangle - y_i \right|^p. 
\]
A popular approach toward minimizing $\mathcal{L}(\beta)$ is the iterative reweighted least square method. At each iteration $t$, given the current solution $\beta_t$, we compute $\Delta\beta_t$ by solving the following reweighted least square problem
\[
    \Delta\beta_t = \mathop{\arg\min}_{\gamma} \sum_{i=1}^n w_{t,i}\left|\langle \beta_t + \Delta\beta_t, x_i\rangle - y_i \right|^2,
\]
where $w_{t,i} = |\langle \beta_t, x_i\rangle - y_i|^{p-2}$. It is clear that $\Delta \beta_{t}$ is given by
\[
    \Delta \beta_t = \left(\sum_{i=1}^n x_{t,i} x_{t,i}^{\top}\right)^{-1}\left(\sum_{i=1}^n (y_i - y_{t,i})x_{t,i} \right),
\]
where $x_{t,i} = \sqrt{w_{t,i}}x_i$ and 
\[
y_{t,i} = \sum_{k=1}^{t}\frac{1}{\sqrt{w_{t,i}}}\langle \Delta\beta_t, x_{t,i}\rangle.
\]
We then update the solution as $\beta_{t+1} = \beta_t + \Delta \beta_t$. The effectiveness of iterative reweighted algorithm is revealed by the theoretical result from \cite{IRA_for_cs/Yinwotao} (Theorem 2.1), where the authors show that $\mathcal{L}(\beta)$ has an unique minimizer and iterative reweighted algorithm converges to the sparse optimal solution. 

We can create a network structure by simply unrolling the iterations of iterative reweighted algorithm and generalizing the functions for updating both feature representation $x$ and regression model $\beta$. As a result, at each the $k$-th layer, we have
\begin{equation}\label{eq_theory_DRNet}
\begin{split}
\delta y_k = y - \sum_{j=1}^{k-1} z_j, \quad \underbrace{x_k = f(x_{k-1}; \delta y_k) + x_{k-1}}_{\mbox{Backcast + Skip Link}},
\\
\quad \Delta\beta_k = g(x_{k}, \delta y_k), \quad \underbrace{z_k = \langle \Delta \beta_k, x_k \rangle}_{\mbox{Forecast Link}},
\end{split}
\end{equation}
where $f(\cdot)$ and $g(\cdot)$ generalizes the above equation for computing $x_{t,i}$ and $\delta \beta_t$. This overall updating scheme through layers leads to what we called double residual link structure, with one link passing the partial prediction result made by individual layer to the final output and one skip link to combine features from two consecutive layers. Given the robustness of iterative reweighted algorithms, we would expect the double residual link structure will enjoy similar robustness in performance. 

\subsection{A Tree-based Model: Feature Selection, Model Ensemble, and Tree based Aggregation}\label{sec_Kolmogorov}

In the previous section, we present a doubly residual link structure that allows us to learn a sparse $\phi(x)$ from training data robustly. However, the nice property of doubly residual link structure may fail when the target function is no longer sparse. To address this problem, we resort to the Kolmogorov–Arnold representation theorem~\cite{hecht1987kolmogorov} (details in Appendix \ref{app:KA_repre}) that any multivariate function can be represented as sums and compositions of univariate functions, i.e., any multivariate function continuous and bounded $h(\boldsymbol{x})$ can be represented as
\[
h(\boldsymbol{x}) = \sum_{k=0}^{2d}\Phi_{k}\left(\sum_{p=1}^{d}\phi _{k,p}(x_{p})\right), 
\]
where both $\Phi_k$ and $\phi_{k,p}$ are univariate functions. It is easy to generalize the above representation theorem by extending $\phi_{k,p}$ into multivariate functions that involves only a small number of input variables. To this end, we can rewrite $h(\boldsymbol{x})$ as
\begin{equation}\label{eq_theory_tree}
h(\boldsymbol{x}) = \underbrace{\sum_{k=0}^{2d} \Phi_k\left(\underbrace{\sum_{p=1}^m \phi_{k,p}\left(\underbrace{\boldsymbol{x}\circ \boldsymbol{m}_{k,p}}_{\mbox{Feature Selection}}\right)}_{\mbox{Model Ensemble}} \right)}_{\mbox{Tree Structure}},
\end{equation}
where $\boldsymbol{m}_{k,p} \in \{0,1\}^d$ is a sparse binary mask and by element-wise product between $\boldsymbol{m}_{k,p}$ with $\boldsymbol{x}$, we will only select a small number of variables from $x$ to be used for $\phi_{k,p}$. Comparing the generalized form with the representation theorem, we can see that each function $\phi_{k,q}$ involves a few more input variables, leading more compact representation (i.e. $m$ will be significantly smaller than $d$). The internal structure $\sum_{p=1}^m \phi_{k,p}\left(\boldsymbol{x} \circ \boldsymbol{m}_{k,p}\right)$ leads to our multi-branch structure, with two major components: feature selection $\boldsymbol{x}\circ \boldsymbol{m}_{k,p}$ and model ensemble $\sum_{p=1}^m \phi_{k,p}(\cdot)$. Finally, to account for the presence of $\Phi_k$, we introduce a tree structure to effectively aggregate information together.

\section{Network Architecture}

\begin{figure}[t!]
  \centering
  \includegraphics[width=1\linewidth]{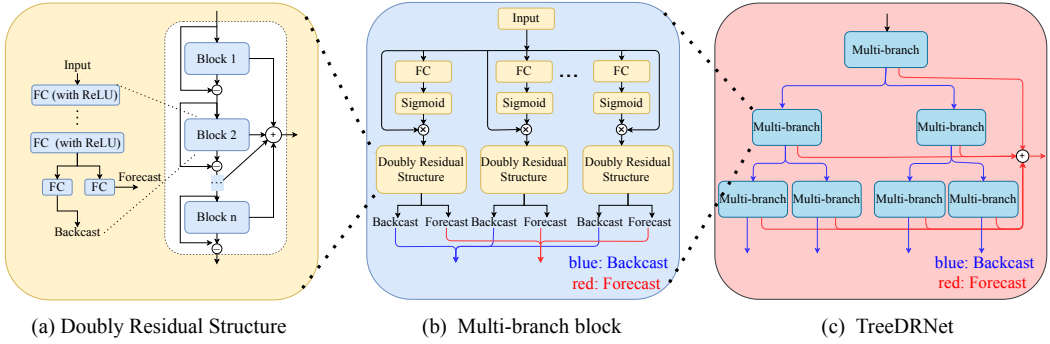}
  \caption{The overall TreeDRNet structure. (a) shows the doubly residual structure, motivated by the robust iterative reweighted algorithm as in Eq.~\eqref{eq_theory_DRNet}. (b) shows the multi-branch block with feature selection (through gating mechanisms), doubly residual structure, and model ensemble (through multiple branch), motivated by the Kolmogorov–Arnold theorem as in Eq.~\eqref{eq_theory_tree}. (c) shows the high-level architecture of the TreeDRNet model with binary-tree structure, where each node of the tree is a multi-branch block with feature selection.}
  \label{fig_TreeDRNet}
\end{figure}

In this section, we design a robust tree-structured doubly residual network (TreeDRNet), as shown in Fig.~\ref{fig_TreeDRNet}, based on the insights of robust iterative reweighted algorithm and Kolmogorov–Arnold representation theorem as analyzed in Section~\ref{sec_motivation_theory}. 
In the following parts, we will elaborate on the TreeDRNet model from low level to high level in detail.
 
\subsection{Doubly Residual Structure}


One of the essential building components in TreeDRNet is the doubly residual structure, as shown in Fig. \ref{fig_TreeDRNet} (a). Note that the residual structure is widely applied in deep learning since ResNet was proposed~\cite{he2016deep}, and N-BEATS~\cite{oreshkin2019n} model also introduces a doubly residual mechanism for solving the univariate time series forecasting problem. 

Different from existing works which adopt residual structure based on the rationale to mitigate the well-known vanishing gradient problem, our doubly residual structure is theoretically motivated by the generalization of the iterative reweighted algorithm in robust regression as discussed in Eq.~\eqref{eq_theory_DRNet}.



Specifically, the doubly residual structure consists of multiple blocks with doubly residual connection, where each block has two parts. The first part is $n$ fully connected layers with ReLU non-linearity for extracting features as
\begin{align}
    x^\ell_{i+1} = ReLU(FC_i(x^\ell_i)), i=0,\cdots,n-1.
\end{align}
The second part consists of two fully connected layers: backcast that projects the features to the same length as the input time horizon $x$ and forecast that projects the features to the same length as the prediction time horizon $y$ as
\begin{align}
    \hat{x}^\ell = FC_b(x^\ell_n),\\
    \hat{y}^{\ell} = FC_f(x^\ell_n).
\end{align}
It's a sequential doubly residual topology, one running over forecast prediction of each layer and the other running over the backcast branch of each layer. Its operation is described by the following equations:
\begin{align}
    x^{\ell+1}=x^{\ell}-\hat{x}^\ell,\\
    y^{\ell+1} = y^{\ell}+\hat{y}^{\ell}.
\end{align}
 The detailed procedure is summarized in Algorithm~\ref{alg:algorithm1} of Appendix~\ref{app:alg}. The doubly residual structure (DRes) is equivalent to continuous residual iteration, reducing residuals through the stacking of multiple blocks and improving the fitting ability of the network.

\subsection{Multi-branch Architecture}
Using fully connected (FC) layers in the doubly residual structure allows simple yet efficient computation procedures. However, compared with more complex structures, such as RNN and CNN, the FC layer's representation power can be relatively weak and potentially incomplete for information acquisition. When dealing with noisy real-world data, representation errors may accumulate through the network, harming the final performance.


Motivated by the Kolmogorov–Arnold theorem as discussed in Eq.~\eqref{eq_theory_tree}, we design a multi-branch architecture as shown in Fig. \ref{fig_TreeDRNet} (b) and the detailed procedure is summarized in Algorithm~\ref{alg:algorithm2} of Appendix~\ref{app:alg}. The multi-branch block integrates the doubly residual structure with feature selection (through gating mechanisms) and model ensemble (through multiple branches) to address the issue mentioned above. In particular,
the term $\boldsymbol{x}\circ \boldsymbol{m}_{k,p}$ and $\sum_{p=1}^m \phi_{k,p}(\cdot)$ in Eq. (2) can be viewed as feature selection and model ensemble respectively.


The main idea here is to perform feature representation through multi-branches and decompose the input information into several parts through the gating mechanism. The different portions of inputs will go through the corresponding branch to obtain different feature representations, and at last, all the branches will be integrated to generate a more compact representation due to efficiency consideration.

The specific formulation of the multi-branch structure is described as follows. Let $$X=[x_1,x_2,\cdots,x_n]$$ denote the input vector. Assuming there are $n$ branches in the module and each branch has a different input $X_i$ produced by the gating mechanisms:
\begin{equation}
    X_i=X\circ\sigma(f_i(X))
\end{equation}
where $\sigma(\cdot)$ is the sigmoid activation function, $f_i$ is $m$ fully connected layers with ReLU non-linearity and $m$ is hyperparameters, $\circ$ is element-wise multiplication. The doubly residual structure (DRes) reads
\begin{align}
    \textrm{F}_i,\textrm{B}_i = DRes_i(X_i)
\end{align}
For $n$ branches, we have $n$ $Backcast$ and $Forecast$. By integrating over all branches, the block forecast ($fc$) and backcast ($bc$) can be defined as the average of the all corresponding branches as\\
\begin{align}
    fc = \frac{1}{n}\sum_{i=1}^{n}\textrm{F}_i, \quad
    bc = \frac{1}{n}\sum_{i=1}^{n}\textrm{B}_i
\end{align}

In order to ensure the computational efficiency of the model, the structure of each branch is set to be the same, which is conducive to the parallel calculation of the model. 
We then use the multi-branch structure as the main module of the TreeDRNet model, and generate the final prediction model by stacking the modules.

\subsection{Tree Structured Doubly Residual Neural Network}
To boost the performance, we further stack multiple multi-branch blocks together to obtain the final TreeDRNet architecture as shown in Fig. \ref{fig_TreeDRNet} (c). This stacking also corresponds to the function $\Phi_k$ in Eq.~\eqref{eq_theory_tree} discussed in Section~\ref{sec_Kolmogorov}.
To prevent the potential information loss, we avoid directly stacking the multi-branch structure to generate a tiled network. Instead, a more effective model generation method is considered in this work. The overall model is built by stacking the above multi-branch structure by a structured tree topology, the specific network structure is given in Fig. \ref{fig_TreeDRNet} and the specific algorithm is summarized in Algorithm~\ref{alg:algorithm3} of Appendix~\ref{app:alg}. Besides the conventional sequential data flow, we use a tree-based data flow in the model where each rectangle represents the multi-branch structure mentioned above. After the previous multi-branch block, the data is passed to the subsequent multi-branch block through two paths like a binary tree. Such structure can further boost the feature generation ability of our model on top of a multi-branch block. It is also expected that our tree-based topology has a stronger regulation ability compared to those of sequential ones because the newly generated features play less role in lower layers (they can only contribute to a single leaf output and their descendants). The model's leaf additive forecast outputs flow increases by almost twice with the one additional depth.  


The process is summarized from a top-down forward path point of view below. A forecast and a backcast can be first obtained through a single multi-branch block, then down to two multi-branch leaf blocks. A multilayer tree-structured model can be recursively constructed. Fig. \ref{fig_TreeDRNet} (c) shows a three-layer structure. In an N-layer model, the $\textrm{i-th}$ ($0<i<=N$) layer has $2^{i-1}$ multi-branch block and outputs $2^{i-1}$  bc and fc. All the bc in this layer work as the root node of the next layer and each will be split into two leafs, i.e. $backcast_i^j$ represents the $\textrm{j-th}$ multi-branch leaf in the $\textrm{i-th}$ layer and in the next layer, we will have $\textrm{bc}_{i+1}^{2j-1}$ and $\textrm{bc}_{i+1}^{2j}$ from this node.

We take the average of all forecasts as predicted value in this layer.
\begin{equation}
    \textrm{fc}_i = \frac{1}{2^{i-1}}\sum_{j=1}^{2^{i-1}}\textrm{fc}_i^j
\end{equation}
 The $forecast_i$ is $i-th$ layer prediction. After running over all blocks of the model, the forecasts of each layer are summed as the model's final prediction.
\begin{equation}
    \textrm{precict} = \sum_{i=1}^L\textrm{fc}_i
\end{equation}

\subsection{Transformation of Covariate Variables}
This section extends our TreeDRNet to the scenario of multi-covariates univariate time series forecasting. Since the correlations between covariate variables and their relationship with the target variable are unknown in this scenario, indiscriminate incorporation of the covariate variables into the model can lead to information redundancy. Therefore, it is vital to quantify the impact of covariate variables on the target variables. Besides specifying the dependence of the target variable on other variables, variable selection can also reduce noise and irrelevant information. We adopt one-dimension convolution to transform covariate variables as
\begin{align} 
& x = Conv(W_x,\hat{X}),
\end{align}
where $\hat{X}\in \mathbb{R}^{t\times d}$ denotes the covariate variables, $W_x$ denotes a one-dimension convolution filter with a filter size of 1,
and $Conv$ is the convolution operation. It learns a set of weights at each time step, indicating the importance of different covariates to the target variable.

\section{Empirical Results and Discussions}\vspace{-2mm}
In this section, we evaluate the proposed TreeDRNet with other state-of-the-art forecasting models, conduct ablation studies, and discuss insights to demonstrate the superior performance of the TreeDRNet model.  More details about baseline models, datasets, implementations, as well as additional experiments are described in Appendix.

\subsection{Main Results}\vspace{-2mm}
For \textbf{multivariate forecasting}, TreeDRNet achieves the best performance on all six benchmark datasets at all horizons as in Table~\ref{tab:multi-benchmarks}. Compared with the SOTA Transformer model Autoformer~\cite{Autoformer}, the proposed TreeDRnet yields an overall \textbf{23.3\%} relative MSE reduction. It is worth noting that for some of the datasets, such as Exchange and Weather, the improvement is even more significant (over \textbf{$30\%$}). Note that the Exchange dataset does not exhibit clear periodicity in its time series, but TreeDRNet can still achieve superior performance. On ETT full benchmarks as shown in Table~\ref{tab:multi-benchmarks-ett} of Appendix~\ref{app:multi_ett}, the proposed TreeDRnet yields an overall \textbf{23.6\%} relative MSE reduction and demonstrates consistency across varying scales since ETTm and ETTh datasets come from same sequence with different sampling rate. 
Overall, the improvement made by TreeDRNet is consistent with varying horizons, implying its superior strength in long term forecasting.  

We also test two typical datasets in the scenario of \textbf{univariate forecasting} and the results are summarized in Table \ref{tab:uni-benchmarks-large} of Appendix~\ref{app:single_full}. Compared with Autoformer, TreeDRNet yields an overall \textbf{54.9\%} relative MSE reduction in a Exchange dataset which doesn't show clear periodicity and a near stationary ETTm2 dataset.
Since our TreeDRNet is a univariate based model, multivariate forecasting and univariate forecasting are intrinsically the same. It is worth to mention that TreeDRNet clearly outperforms N-BEATS~\cite{oreshkin2019n}, another MLP based univariate baseline. 

Note that we treat the above datasets as forecasting without covariates. To extend and validate our model's strength in in the scenario of \textbf{multicovariates forecasting}, we test it on two large dataset: Vol and Retail as in Table~\ref{table_covariates_result} of Appendix~\ref{app:Multicovariates}. Compared with the SOTA Transformer model TFT~\cite{lim2021temporal}, we can obtain a comparable result with incremental improvement. It might indicates attention mechanism in Transformer have a strong ability in utilizing covariates information. Some future work can be done in integrate Transformer with our TreeDRNet model.

\vskip -0.2in
\begin{table*}[t]
\centering
\caption{multivariate long-term series forecasting results on six datasets with input different input length based on datasets and prediction length $O \in \{96,192,336,720\}$ (For ILI dataset, we set prediction length $O \in \{24,36,48,60\}$). A lower MSE indicates better performance.}
\scalebox{0.85}{
\begin{tabular}{c|c|cccccccccccccccc}
\toprule
\multicolumn{2}{c|}{Methods}&\multicolumn{2}{c|}{TreeDRNet}&\multicolumn{2}{c|}{Autoformer}&\multicolumn{2}{c|}{TCN}&\multicolumn{2}{c|}{Informer}&\multicolumn{2}{c|}{LogTrans}&\multicolumn{2}{c}{Reformer}\\
\midrule
\multicolumn{2}{c|}{Metric} & MSE  & MAE &MSE  & MAE& MSE  & MAE& MSE  & MAE& MSE  & MAE& MSE  & MAE\\
\midrule
\multirow{4}{*}{\rotatebox{90}{$ETTm2$}} &96 & \textbf{0.179} & \textbf{0.245} &0.255  &0.339 &0.258 &0.359 &0.365  &0.453  &0.768  &0.642  &0.658  &0.619    \\
                        & 192 & \textbf{0.233} & \textbf{0.309}   &0.281 &0.340 &0.396 &0.452 &0.533  &0.563  &0.989  &0.757  &1.078  &0.827    \\
                        & 336 & \textbf{0.303} & \textbf{0.356}   &0.339  &0.372 &0.545 &0.542 &1.363&0.887  &1.334  &0.872  &1.549  &0.972     \\
                        & 720 & \textbf{0.387} & \textbf{0.413}   &0.422  &0.419 &0.916 &0.731 &3.379  &1.338 & 3.048 &1.328  &2.631  &1.242      \\
\midrule
\multirow{4}{*}{\rotatebox{90}{$Electricity$}} &96  &\textbf{0.163}  &\textbf{0.267}   &0.201  &0.317 &0.260 &0.355 &0.274  &0.368  &0.258  &0.357  &0.312  &0.402    \\
                                               & 192 &\textbf{0.185} & \textbf{0.289}  &0.222  &0.334 &0.275 &0.369 &0.296 &0.386  &0.266 &0.368  &0.348  &0.433    \\
                                                & 336 & \textbf{0.203} & \textbf{0.310} &0.231 &0.338 &0.288 &0.381 &0.300  &0.394  &0.280 &0.380  &0.350  & 0.433    \\
                                                & 720 & \textbf{0.236} & \textbf{0.238} &0.339 &0.361 &0.325 &0.410 &0.373  &0.439 &0.283  &0.376  &0.340  &0.420     \\
\midrule
\multirow{4}{*}{\rotatebox{90}{$Exchange$}} &96  & \textbf{0.091} & \textbf{0.224} &0.197  &0.323 &0.357 &:0.400 &0.847  &0.752  &0.968  &0.812  &1.065  &0.829    \\
                                            & 192 & \textbf{0.219} & \textbf{0.346} &0.300  &0.369 &3.048 &0.552 &0.518   &0.895 &1.040  &0.851  &1.188  & 0.906   \\
                                            & 336 & \textbf{0.412} & \textbf{0.502} &0.509  &0.524 &0.859 &0.674 &1.672  &1.036  &1.659  &1.081  &1.357  &0.976     \\
                                            & 720 & \textbf{0.690} & \textbf{0.647} &1.447  &0.941 &1.487 &0.935 &2.478  &1.310  &1.941  &1.127  &1.510  &1.016     \\
\midrule
\multirow{4}{*}{\rotatebox{90}{$Traffic$}} &96  &\textbf{0.417}& \textbf{0.287}    &0.613  &0.388 &0.872 &0.512 &0.719  &0.391  &0.684  &0.384  &0.732  &0.423    \\
                                           & 192 & \textbf{0.433} & \textbf{0.299} &0.616. &0.382 &0.896 &0.528 &0.696 &0.379  &0.685  &0.390  &0.733  &0.420    \\
                                           & 336 & \textbf{0.451} & \textbf{0.306} &0.622  &0.337 &0.918 &0.535 &0.777  &0.420  &0.733  &0.408  &0.742  &0.420     \\
                                           & 720 & \textbf{0.518} &\textbf{0.335}  &0.660  &0.408 &0.964 &0.549 &0.864  &0.472  &0.717  &0.396  &0.755  &0.423     \\
\midrule
\multirow{4}{*}{\rotatebox{90}{$Weather$}} & 96 & \textbf{0.156} & \textbf{0.203}.  &0.266  &0.336 &0.190 &0.248 &0.300  &0.384  &0.458  &0.490  &0.689  &0.596    \\
                                           & 192 & \textbf{0.211} & \textbf{0.256}  &0.307  &0.367 &0.629 &0.232 &0.288  &0.544  &0.658  &0.589  &0.752  &0.638    \\
                                           & 336 &\textbf{0.266} &\textbf{0.317}    &0.359  &0.395 &0.277 &0.325 &0.578  &0.523  &0.797  &0.652  &0.639  &0.596    \\
                                           & 720 &0.365 & 0.364.  &0.578  &0.578 &\textbf{0.345} &\textbf{0.377}  &1.059  &0.741  &0.869  &0.675  &1.130  &0.792    \\
\midrule
\multirow{4}{*}{\rotatebox{90}{$ILI$}} & 24 & \textbf{2.394} & \textbf{1.019}  &3.483  &1.287 &5.968 &1.741 &5.764  &1.677  &4.480  &1.444  &4.400 &1.382    \\
                                       & 36 & \textbf{2.228} & \textbf{0.989}  &3.103  &1.148 &6.858 &1.879 &4.755  &1.467  &4.799  &1.467  &4.783  &1.448    \\
                                       & 48 & \textbf{2.443} & \textbf{1.034}  &2.669  &1.085 &6.968 &1.892 &4.763  &1.469  &4.800  &1.468  &4.832  &1.465    \\
                                       & 60 & \textbf{2.938} & \textbf{1.150}  &2.770  &1.125 &7.127 &1.918 &5.264  &1.564  &5.278  &1.560  &4.882  &1.483    \\
\bottomrule
\end{tabular}
\label{tab:multi-benchmarks}
}
\end{table*}

\begin{table*}[t]

\centering
\caption{Efficiency experiment: 96 to 720 experiment for ETTm2 dataset, the reported time is per-sample training \& inference time.}\vspace{-1mm}
\scalebox{1.00}{
\begin{tabular}{c|c|ccccccccc}
\toprule
Model & TreeDRNet & Autoformer & Informer & Reformer & Transformer&TCN\\
\midrule


Training Time (ms) & 0.156 & 2.302 & 2.494 & 1.752 & 3.403&0.304\\
Inference Time (ms) & 0.110 & 1.534 & 2.108  & 1.154 & 2.477 &0.210\\
\bottomrule
\end{tabular}
}
\label{tab:runningtime}
\vskip -0.1in
\end{table*}
\begin{table*}[t]
\centering
\caption{The robustness experiment of forecasting 720 time steps task for ETTm2 dataset with prolonging input length. Each experiment is done five times and reports the mean}\vspace{-1mm}
\scalebox{1.0}{
\begin{tabular}{c|c|ccccccccc}
\toprule
\multicolumn{2}{c|}{Model} & TreeDRNet & Autoformer & Informer\\
\midrule

\multirow{6}{*}{\rotatebox{90}{$Input\ Length$}}
& 96 & 0.414 & \textbf{0.433} & \textbf{3.379}\\
&192 & 0.412 & 0.463 & 4.907\\
&336 & 0.409 & 0.545 & 5.082\\
&720 & 0.407 & 0.600 & 6.893\\
&1440& 0.388 & OOM & OOM \\
&3600& \textbf{0.387} & OOM & OOM\\



\bottomrule
\end{tabular}
}
\label{tab:robustness}
\vskip -0.1in
\end{table*}

\subsection{Efficiency Analysis}
Although Transformer-based methods have significantly improved state-of-the-art results for time series forecasting, they usually have high computational complexity and expensive memory for long sequences. 
As shown in Table~\ref{tab:runningtime}, the training and inference speed of the TreeDRNet is over \textbf{10 times} faster compared to various Transformer-based models. It is worth mentioning that TCN also has a short per-sample training/inference time compared to Transformer base models. However, TreeDRNet is still 2 times faster and much more accuracy compared to TCN as shown in Table~\ref{tab:multi-benchmarks} and Table~\ref{tab:runningtime}. So our contribution is not simply to achieve faster speed but achieve that with a SOTA performance. 
Extra experiments about memory usage are sumarized in Fig.~\ref{fig_memory} of Appendix~\ref{app:mem}, where the results show that our MLP-based TreeDRNet model have a better performance with much less memory usage and higher efficiency. 
The memory usage of our TreeDRNet model is two orders of magnitude smaller than its various Transformer counter parts. And our memory usage increases \textbf{linearly} with prolonging input length compared to $O(nlogn)$ increasing rate of Transformer based models. 


\subsection{Robustness Analysis}


\paragraph{Robustness with prolonged input sequence}
We perform experiments to show the robustness of our model compared to recent dominating Transformer-based models. 
TreeDRNet show a robust forecasting with prolonged input sequence as shown in Figure \ref{Robutness}.

Specifically, as shown in Table \ref{tab:robustness}, when we compare the prediction results with prolonged input length, Transformer-based models deteriorate quickly with prolonging input length. This phenomenon makes a lot of carefully designed Transformers impractical in long-term forecasting tasks since they can't effectively utilize long input's information.
On the other hand, our proposed TreeDRNet enjoys robust ability to not only handle but utilize the prolonged noise in real life input signal. This experiment raises the importance of \textbf{robustness} in long-term time series forecasting. The ability to process long input is far from enough, how to robustly extract more information from long input sequence to boost forecasting performance should be the key question.

\paragraph{Robustness with Attacks}
Extended noisy historical data is one way to test our model's robustness. Explicitly injecting in noise or noise attack\cite{hsieh-etal-2019-robustness} is another way. As shown in Appendix, we conducted a set of noisy injection attack experiments to show our model's robustness. 

\paragraph{Robustness with Noise Injection}
Extended noisy historical data is one way to test the model's robustness. Explicitly injecting in noise is another way. As shown in Appendix, we conducted three noisy injection attack experiments to show its robustness compared to transformer-based sota model.

\paragraph{Robustness theoretical Analysis: attention vs double residual like structure}
The heavy vanilla attention module with small hidden dimension and layer number can be less sensitive to noise theoretically as shown in Appendix. 

\subsection{Other Experiments}
Other experiments are provided in Appendix. TreeDRNet is designed based on Kolmogorov–Arnold representation theorem with feature selection, model ensemble, and tree structure. An ablation analysis is conducted to show that each component plays an essential role in the model as in Appendix~\ref{app:ablation}. Moreover, we conduct sensitivity experiments to show the hidden layer dimension, the number of multi-branch block and depth of the tree's effect for TreeDRNet as in Appendix~\ref{app:Sensitivity}. More experiments are conducted to show the effectiveness of double residual linked structure, which can be found in Appendix.



\section{Conclusion}
Here, we introduce a robust long-term prediction deep model named TreeDRNet, inspired by the iterative reweighted algorithm and Kolmogorov–Arnold representation theorem. The TreeDRNet combines doubly residual links, feature selection, model ensemble, and a tree structure to boost performance. Extensive experiments over multiple benchmark datasets across diverse domains show that the proposed TreeDRNet model enjoys the advantage of high computational efficiency compared to Transformer-based forecasting models and improves the performance of state-of-the-art results significantly by reducing prediction errors 20\% to 40\% for long-term multivariate time series forecasting.

\newpage


\bibliography{example_paper}
\bibliographystyle{neurips_2022}

\newpage
\appendix

\section{Appendix}

\appendix

\section{Iteratively Reweighted Least Square}
\label{app:iter_rewei}

The method of iteratively reweighted least squares (IRLS) is a powerful optimization algorithm used to solve an $L_p$ approximation problem \cite{daubechies2010iteratively}. 

The IRLS algorithm is an iterative method and the main process is presented as follows:

To find the parameters $\beta = (\beta_1, \ldots, \beta_k)^\top$ which minimize the $L_p$ norm for the linear regression problem:
\begin{equation}
{  {\underset {\boldsymbol {\beta }}{\operatorname {arg\,min} }}{\big \|}\mathbf {y} -X{\boldsymbol {\beta }}\|_{p}^p = {\underset {\boldsymbol {\beta }}{\operatorname {arg\,min} }}\sum _{i=1}^{n}\left|y_{i}-X_{i}{\boldsymbol {\beta }}\right|^{p},}
\end{equation}
the IRLS algorithm at step $t + 1$ involves solving the weighted linear least squares problem:

\begin{equation}
{\boldsymbol {\beta }}^{(t+1)}={\underset {\boldsymbol {\beta }}{\operatorname {arg\,min} }}\sum _{i=1}^{n}w_{i}^{(t)}\left|y_{i}-X_{i}
{\boldsymbol {\beta }}\right|^{2}=(X^{\rm {T}}W^{(t)}X)^{-1}X^{\rm {T}}W^{(t)}\mathbf {y}
\end{equation}
where $W^{(t)}$ is the diagonal matrix of weights and updated after each iteration to:
\begin{equation}
{  w_{i}^{(t)}={\big |}y_{i}-X_{i}{\boldsymbol {\beta }}^{(t)}{\big |}^{p-2}}
\end{equation}
In the case $p = 1$, this corresponds to least absolute deviation regression (in this case, the problem would be better approached by use of linear programming methods, so the result would be exact) and the formula is:
\begin{equation}
{  w_{i}^{(t)}={\frac {1}{{\big |}y_{i}-X_{i}{\boldsymbol {\beta }}^{(t)}{\big |}}}.}
\end{equation}
The double residual link structure can be interpreted as iterative reweighted algorithms, thus they share the similar characteristic of robustness.

\section{Kolmogorov-Arnold representation theorem}
\label{app:KA_repre}
The Kolmogorov-Arnold representation theorem~\cite{hecht1987kolmogorov} states that for any continuous function $f:[0,1]^d \rightarrow \R$, there exist univariate continuous functions $\Phi_q$, $\phi_{p,q}$ such that
\begin{equation}
    f(x_1, \ldots , x_d) =  \sum_{q=0}^{2d} \Phi_q(\sum^d_{p=1} \phi_{p,q}(x_p))
\end{equation}

This means the an exact representation of d-variate function can be constructed with the sum of (2d+1)(d+1) univariate functions. The Kolmogorov-Arnold representation is used to explain the effectiveness of a two-layer neural networks because their similar structure. A two-layer feedforward neural network with layer width $m_1$ and $m_2$ can be expressed as:
\begin{equation}
    f(\bm{x})=\sum_{q=1}^{m_1} d_q \sigma(\sum_{p=1}^{m_2} b_{pq} \sigma (\bm{w}_p^\top \bm{x} + a_p) +c_q)    
\end{equation}
where $\sigma$ is the activation function, and $\bm{w}_p$, $a_p$, $b_{pq}$, $c_q$ and $d_q$ are parameters. 

In the proposed TreeDRNet, we use multi-branch ensemble in place of $\phi_{k,p}$,subse and tree based aggregation in place of $\Phi_k$. Either of them shows stronger expressive power than a single feedforward layer, which leads to a better performance of the proposed framework.

\newpage
\section{Algorithm} \label{app:alg}
\subsection{The algorithm of Doubly Residual Structure (DRes)}
\begin{algorithm}[h]
\caption{The algorithm of Doubly Residual Structure (DRes)}
\label{alg:algorithm1}
\begin{algorithmic}[1] 
\STATE \textbf{Input:} $x,y=0$\\
\STATE Let $x^1=x, y^1=0$.
\FOR {$\ell=0:L-1$}
    \STATE $x^{\ell,0} = x^{\ell}$
    \FOR {$i=0:n-1$} 
        \STATE $x^{\ell}_{i+1}=ReLU(FC^\ell_i((x^{\ell}_{i}))$
    \ENDFOR  
    \STATE $\hat{x}^\ell=FC^\ell_b(x^{\ell}_n)$
    \STATE $\hat{y}^\ell=FC^\ell_f(x^{\ell}_n)$
    \STATE $x^{\ell+1}=x^{\ell}-\hat{x}^\ell$
    \STATE $y^{\ell+1} = y^{\ell}+\hat{y}^{\ell}$
\ENDFOR
\STATE $\textrm{B}=x^L$, $\textrm{F}=y^L$
\STATE \textbf{return} $\textrm{B}$,$\textrm{F}$
\end{algorithmic}
\end{algorithm}

\subsection{The algorithm of Multi-branch block}
\begin{algorithm}[h]
\caption{The algorithm of Multi-branch block}
\label{alg:algorithm2}
\begin{algorithmic}[1] 
\STATE \textbf{Input:} input data $X$ and the number of branches $m$.\\
\FOR {$\ell = 1$ to $m$}
    \STATE $X_i=X\circ\sigma(f_i(X))$
    \STATE $\textrm{B}_i, \textrm{F}_i = DRes_i(X_i)$
\ENDFOR
\STATE $bc = \frac{1}{m}\sum_{i=1}^{m}\textrm{B}_i$
\STATE $fc = \frac{1}{m}\sum_{i=1}^{m}\textrm{F}_i$
\STATE \textbf{return} $\textrm{Backcast}$, $\textrm{Forecast}$
\end{algorithmic}
\end{algorithm}

\subsection{The algorithm of TreeDRNet}
\begin{algorithm}[h]
\caption{The algorithm of TreeDRNet}
\label{alg:algorithm3}
\begin{algorithmic}[1] 
\STATE \textbf{Input:} $bc,~~fc=0$\\
\STATE $bc_1^1,fc_1^1= \textrm{Multi-branch}_1^1(bc)$.
\FOR {$i=2:L$}
  \FOR {$j=1:2^{i-1}$} 
    \STATE $bc_{i}^{j},fc_{i}^{j}=\textrm{Multi-branch}_i^j(bc_{i-1}^{\lceil \frac{j}{2} \rceil})$\\
  \ENDFOR  
  \STATE $fc_i$ = $\frac{1}{2^{i-1}}\sum_{j=1}^{2^{i-1}}fc_i^j$
\ENDFOR
\STATE $\textrm{precict} = \sum_{i=1}^Lfc_i$
\STATE \textbf{return} $\textrm{predict}$
\end{algorithmic}
\end{algorithm}

\section{Datasets}

We consider eight benchmark datasets across diverse domains.
The details of the experiment datasets are summarized as follows: 1) ETT \cite{haoyietal-informer-2021} dataset contains two sub-dataset: ETT1 and ETT2, collected from two electricity Transformers at two stations. Each of them has two versions in different resolutions (15min \& 1h). ETT dataset contains multiple series of loads and one series of oil temperatures. 2) Electricity\footnote{https://archive.ics.uci.edu/ml/datasets/ElectricityLoadDiagrams 20112014} dataset contains the electricity consumption of clients with each column corresponding to one client. 3) Exchange \cite{lai2018modeling} contains the current exchange of 8 countries. 4) Traffic\footnote{http://pems.dot.ca.gov} dataset contains the occupation rate of freeway system across the State of California. 5) Weather\footnote{https://www.bgc-jena.mpg.de/wetter/} dataset contains 21 meteorological indicators for a range of 1 year in Germany. 6) Illness\footnote{https://gis.cdc.gov/grasp/fluview/fluportaldashboard.html} dataset contains the influenza-like illness patients in the United States. 7) Retail: Favorita Grocery Sales Dataset from the Kaggle competition \cite{calero2018corporacion} that combines metadata for different products and the stores. 8) Volatility (or Vol.): The OMI realized library contains daily realized volatility values of 31 stock indices computed from intraday data, along with their daily returns. For our experiments, we consider forecasting over the next week (i.e. 5 business days) using the information over the past year (i.e. 252 business days).
Note that the first six datasets are used for multivariate forecasting while the last two datasets are used for  multicovariates univariate time series forecasting.
Table \ref{tab:dataset} summarizes feature details (Sequence Length: Len, Dimension: Dim, Frequency: Freq) of the eight datasets. All datasets are split into the training set, validation set and test set by the ratio of 7:1:2. 

\begin{table}[t]
\begin{subtable}{\textwidth}
\caption{Summarized feature details of eight datasets.}\vspace{-3mm}
\label{tab:dataset}
\begin{center}
\begin{sc}
\scalebox{0.95}{
\begin{tabular}{l|cccr}
\toprule
Dataset & len & dim & freq \\
\midrule
ETTm2 & 69680 & 8 & 15 min\\
Electricity & 26304 & 322 & 1h & \\
Exchange & 7588 & 9 & 1 day\\
Traffic & 17544 & 863 & 1h & \\
Weather & 52696 & 22 & 10 min & \\
ILI & 966 & 8 & 7 days\\
Vol. &86582  & 22 & 1 day\\
Retail &58106641  &22  & 1 day\\
\bottomrule
\end{tabular}
}
\end{sc}
\end{center}
\vskip -0.1in
\end{subtable}
\end{table}

\section{Baselines}


Since classic models like ARIMA and basic RNN/CNN models perform relatively inferior as shown in \cite{haoyietal-informer-2021} and~\cite{Autoformer}, we mainly include four state-of-the-art (SOTA) Transformer-based models for comparison, i.e., Autoformer~\cite{Autoformer}, Informer~\cite{haoyietal-informer-2021}, LogTrans~\cite{Log-transformer-shiyang-2019} and Reformer~\cite{DBLP:conf/iclr/KitaevKL20-reformer}, convolution and recurrent networks TCN (\cite{bai2018empirical}) as baseline models. 
Note that since Autoformer holds the best performance in all the six multivariate benchmarks, it is used as the main baseline model for comparison.  

For multicovarates experiment we compare the performance of TreeDRNet with 1) simple sequence-to-sequence models with global contexts (Seq2Seq); and 2) the Multi-horizon Quantile Recurrent Forecaster (MQRNN) \cite{wen2017multi}; 3) DeepAR with LSTM blocks in the model; and 4) a transformer-based architecture TFT~\cite{lim2021temporal} which hold the SOTA performance of the two datasets (Vol and Retail) we adopt. 

\section{Implementation Details}
\label{app:exp:implement}
Our model is trained using ADAM \cite{kingma_adam:_2017} optimizer with a learning rate of $1e^{-4}$. The batch size is set to 32. An early stopping counter is employed to stop the training process after three epochs if no loss degradation on the valid set is observed. The mean square error (MSE) and mean absolute error (MAE) are used as metrics. All experiments are repeated 5 times and the mean of the metrics is used in the final results. All the deep learning networks are implemented in PyTorch \cite{NEURIPS2019_9015_pytorch} and trained on NVIDIA V100 32GB GPUs.

\section{Experimental Details}
Note that for multivariate datasets, we produce
forecast for each feature in the dataset and metrics are averaged across all features. Since our TreeDRNet is a univariate based model, each variable is predicted using only its own history as input. We divide each data set into three parts: training set, validation set and test set - training set for network learning, the validation set for hyperparameter tuning, and the test set for the model performance evaluation.

\section{Univariate Forecasting Full Result}\label{app:single_full}
The univariate forecasting result table is shown in Table \ref{tab:uni-benchmarks-large}

\begin{table*}[h]
\setlength\tabcolsep{3pt} 
\centering
\begin{small}
\caption{Univariate long-term series forecasting results on two datasets with prediction length $O \in \{96,192,336,720\}$. A lower MSE indicates better performance, and the best results are highlighted in bold.}\vspace{-1mm}
\scalebox{0.85}{
\begin{tabular}{c|c|cccc|cccc}
\toprule
\multirow{2}{*}{Methods} & \multirow{2}{*}{Metric} &\multicolumn{4}{c|}{ETT}&\multicolumn{4}{c}{Exchange}\\
& &96&192&336&720&96& 192 & 336 & 720 \\
\midrule
\multirow{2}{*}{TreeDRNet} & MSE & \textbf{0.052} &  \textbf{0.087} &  \textbf{0.112} &  \textbf{0.153} &   \textbf{0.144} &  \textbf{0.234} &  \textbf{0.467} &  \textbf{0.876} \\
& MAE & \textbf{0.125} & \textbf{0.223} &  \textbf{0.246} &  \textbf{0.302} &   \textbf{0.306} &  \textbf{0.358} &  \textbf{0.487} &  \textbf{0.688} \\\midrule

\multirow{2}{*}{Autoformer} & MSE & 0.065 & 0.118 & 0.154 & 0.182 & 0.241 & 0.273 & 0.508 & 0.991  \\
                            & MAE & 0.189 & 0.256 & 0.305 & 0.335 & 0.387 & 0.403 & 0.539 & 0.768 \\\midrule
                            
\multirow{2}{*}{N-BEATS}    & MSE & 0.082 & 0.120 & 0.226 & 0.188 & 0.156 & 0.669 & 0.611 & 1.111  \\
                            & MAE & 0.219 & 0.268 & 0.370 & 0.338 & 0.299 & 0.665 & 0.605 & 0.860 \\\midrule
                            
\multirow{2}{*}{Informer}   & MSE & 0.088 & 0.132 & 0.180 & 0.300 & 0.591 & 1.183 & 1.367 & 1.872 \\
                            & MAE & 0.225 & 0.283 & 0.336 & 0.435 & 0.615 & 0.912 & 0.984 & 1.072 \\\midrule
                            
\multirow{2}{*}{LogTrans}   & MSE &0.082&	0.133&	0.201&	0.268&	0.279& 1.950& 2.438& 2.010 \\
                            & MAE &0.217&	0.284&	0.361&	0.407&	0.441& 1.048& 1.262& 1.247\\\midrule
                            
\multirow{2}{*}{Reformer}   & MSE &  0.131 &  0.186 &  0.220 &  0.267 &   1.327 &  1.258 &  2.179 &  1.280 \\
                            & MAE &   0.288 &  0.354 &  0.381 &  0.430 &   0.944 &  0.924 & 1.296 &  0.953 \\
\bottomrule

\end{tabular}
}
\label{tab:uni-benchmarks-large}
\end{small}
\vskip -0.1in
\end{table*}

\section{Multivariate Forecasting Result for ETT full benchmark}\label{app:multi_ett}
The multivariate forecasting result table for ETT four datasets is shown in Table \ref{tab:multi-benchmarks-ett}

\begin{table*}[h]
\centering
\caption{Multivariate long sequence time-series forecasting results on ETT full benchmark. The best results are highlighted in bold.}\vspace{-1mm}
\begin{small}
\scalebox{0.79}{
\begin{tabular}{c|c|cccccccccccccc}
\toprule
\multicolumn{2}{c|}{Methods}&\multicolumn{2}{c|}{TreeDRNet}&\multicolumn{2}{c|}{Autoformer}&\multicolumn{2}{c|}{Informer}&\multicolumn{2}{c|}{LogTrans}&\multicolumn{2}{c}{Reformer}\\
\midrule
\multicolumn{2}{c|}{Metric} & MSE & MAE& MSE  & MAE& MSE  & MAE& MSE  & MAE& MSE  & MAE\\
\midrule
\multirow{4}{*}{\rotatebox{90}{$ETTh1$}}
& 96  &\textbf{0.427} &\textbf{0.457} & 0.449& 0.459& 0.865 & 0.713& 0.878& 0.740& 0.837& 0.728\\
& 192 &\textbf{0.463} &\textbf{0.485} & 0.500& 0.482& 1.008 & 0.792& 1.037& 0.824& 0.923& 0.766\\
& 336  &\textbf{0.504} &\textbf{0.518} & 0.521& 0.496& 1.107 & 0.809& 1.238& 0.932& 1.097& 0.835\\
& 720 &0.629 &0.591 & \textbf{0.514}& \textbf{0.512}& 1.181 &0.865&  1.135& 0.852& 1.257& 0.889\\
\midrule
\multirow{4}{*}{\rotatebox{90}{$ETTh2$}}
& 96  &\textbf{0.155} &\textbf{0.287} & 0.358& 0.397& 3.755& 1.525& 2.116& 1.197 &2.626 &1.317\\
& 192 &\textbf{0.181} &\textbf{0.301} & 0.456& 0.452& 5.602& 1.931& 4.315& 1.635 &11.12 &2.979\\
& 336 &\textbf{0.259} &\textbf{0.361} & 0.482& 0.486& 4.721& 1.835& 1.124& 1.604 &9.323 &2.769\\
& 720 &\textbf{0.312} &\textbf{0.420} & 0.515& 0.511& 3.647& 1.625& 3.188& 1.540 &3.874 &1.697\\
\midrule
\multirow{4}{*}{\rotatebox{90}{$ETTm1$}}
& 96  &\textbf{0.311} &\textbf{0.350} & 0.505& 0.475& 0.672& 0.571& 0.600& 0.546 &0.538 &0.528\\
& 192 &\textbf{0.381} &\textbf{0.400} & 0.553& 0.496& 0.795& 0.669& 0.837& 0.700 &0.658 &0.592\\
& 336 &\textbf{0.423} &\textbf{0.436} & 0.621& 0.537& 1.212& 0.871& 1.124& 0.832 &0.898 &0.721\\
& 720 &\textbf{0.573} &\textbf{0.524} & 0.671& 0.561& 1.166& 0.823& 1.153& 0.820 &1.102 &0.841\\
\midrule
\multirow{4}{*}{\rotatebox{90}{$ETTm2$}} &96 &\textbf{0.179} &\textbf{0.245} &0.255  &0.339  &0.365  &0.453  &0.768  &0.642  &0.658  &0.619 \\
                        & 192 &\textbf{0.233}  &\textbf{0.309}  &0.281 &0.340 &0.533  &0.563  &0.989  &0.757  &1.078  &0.827 \\
                        & 336  & \textbf{0.303} &\textbf{0.356} &0.339  &0.372  &1.363&0.887  &1.334  &0.872  &1.549  &0.972 \\
                        & 720 &\textbf{0.387} &\textbf{0.412} &0.422  &0.419  &3.379  &1.338 & 3.048 &1.328  &2.631  &1.242 \\
\bottomrule
\end{tabular}
}
\end{small}
\label{tab:multi-benchmarks-ett}
\vskip -0.1in
\end{table*}

\section{Multiple random runs for TreeDRNet}

Table~\ref{tab:std} lists both mean and standard deviation (STD) for TreeDRNet and Autoformer with 5 runs. We observe a quite small variance in the performance of TreeDRNet.

\vspace{-.2cm}
\begin{table}[h]
\label{sample-table-kstest}
\centering
\vskip -0.2in
\caption{A subset of the benchmark showing both Mean and STD.}
\begin{small}

\scalebox{0.75}{
\begin{tabular}{c|c|ccccc}
\toprule
\multicolumn{2}{c|}{MSE}& ETTm2 & Electricity & Exchange & Traffic\\ 
\midrule
\multirow{4}{*}{\rotatebox{90}{TreeDRNet}} 
& 96 & 0.179 $\pm$ 0.006 & 0.163 $\pm$ 0.002 & 0.091 $\pm$ 0.002 & 0.417 $\pm$ 0.007 \\
& 192 & 0.233 $\pm$ 0.016 & 0.185$\pm$ 0.002 & 0.219$\pm$ 0.006 & 0.433 $\pm$ 0.004 \\
& 336 & 0.303 $\pm$ 0.006 & 0.203$\pm$ 0.002 & 0.412$\pm$ 0.013 & 0.451 $\pm$ 0.007 \\
& 720 & 0.387 $\pm$ 0.008 & 0.238$\pm$ 0.002 & 0.690$\pm$ 0.016 & 0.518 $\pm$ 0.005 \\
\midrule
\multirow{4}{*}{\rotatebox{90}{Autoformer}} 
& 96 & 0.255 $\pm$ 0.020 & 0.201$\pm$ 0.003 & 0.197$\pm$ 0.019 & 0.613$\pm$ 0.028 \\
& 192 & 0.281 $\pm$ 0.027 & 0.222$\pm$ 0.003 & 0.300$\pm$ 0.020 & 0.616$\pm$ 0.042 \\
& 336 & 0.339 $\pm$ 0.018 & 0.231$\pm$ 0.006 & 0.509$\pm$ 0.041 & 0.622$\pm$ 0.016 \\
& 720 & 0.422 $\pm$ 0.015 & 0.254$\pm$ 0.007 & 1.447$\pm$ 0.084 & 0.419$\pm$ 0.017 \\
\bottomrule
\end{tabular}
}
\label{tab:std}
\end{small}
\end{table}

\section{Multicovariates Forecasting Experiments}\label{app:Multicovariates}
Table~\ref{table_covariates_result} summarizes the multicovariates univariate forecasting results of P50 and P90 quantile losses on both Vol and Retail datasets.
\begin{table*}[h]
  \caption{Multi-covariates univariate forecasting results on Vol and Retail datasets. Performance metrics are based on P50 and P90 quantile losses , where percentages in brackets reflect the increase/decrease in quantile loss versus TreeDRNet (lower q-Risk indicates better performance). The best results are high-lighted in bold.}\vspace{-1mm}
  
  \label{table_covariates_result}
  \begin{center}

   \begin{small}
  \scalebox{0.80}{
  \begin{tabular}{ccccccc}
    \toprule
                          & DeepAR   & ConvTrans   & Seq2Seq & MGRNN & TFT   & TreeDRNet\\
    \midrule
    \texttt{Vol}           & 0.050(+22\%)    & 0.047(+17\%)       &  0.042(+7\%)  & 0.042(+7\%)   & \textbf{0.039(+0\%)} & \textbf{0.039} \\
    \texttt{Retail}        & 0.574(+39\%)    & 0.429(+19\%)       &  0.411(+16\%)  & 0.379(+8\%)  & 0.354(+2\%) & \textbf{0.347} \\
    \bottomrule
    \multicolumn{6}{c}{(a) P50 losses on popular datasets.}
    \\
  \end{tabular}
  }\vspace{2mm}
  \scalebox{0.80}{
    \begin{tabular}{ccccccc}
    \toprule
                          & DeepAR          & ConvTrans       & Seq2Seq        & MGRNN          & TFT         & TreeDRNet\\
    \midrule
    \texttt{Vol}           & 0.024(+21\%)    & 0.024(+21\%)   &  0.021(+10\%)  & 0.021(+10\%)   & 0.020(+5\%) & \textbf{0.019} \\
    \texttt{Retail}        & 0.230(+36\%)    & 0.192(+24\%)   &  0.157(+7\%)   & 0.152(+4\%)    & 0.147(+1\%) & \textbf{0.146} \\
    \bottomrule
    \multicolumn{6}{c}{(b) P90 losses on popular datasets.}
  \end{tabular}
  }
  \end{small}
  \end{center}
\end{table*}









\section{Ablation Analysis}\label{app:ablation}
TreeDRNet is designed based on Kolmogorov–Arnold representation theorem with feature selection, model ensemble, and tree structure. We believe each component plays an important role in TreeDRNet. To quantify the benefits from each ingredient of our proposed architecture, we perform the ablation analysis, i.e. removing each component from the network to test performance. For simplicity, we define the following set of ablated models: 

\begin{itemize}
    \item TreeDRNet$_{woF}$: without feature selection block: each branch has the same input.
    \item TreeDRNet$_{woE}$: without model ensemble: only have one branch in each block.
    \item TreeDRNet$_{woT}$: without tree structure: the network is sequential structure.
\end{itemize}
\begin{table}[h]
\centering
\caption{Component ablation experients. The best results are highlighted in bold.} 
\scalebox{0.85}{
\begin{tabular}{c|c|cccccccccccc}
\toprule
\multicolumn{2}{c|}{Model}&\multicolumn{2}{c|}{TreeDRNet}&\multicolumn{2}{c|}{TreeDRNet$_{woF}$}&\multicolumn{2}{c|}{TreeDRNet$_{woE}$}&\multicolumn{2}{c}{TreeDRNet$_{woT}$}\\
\midrule
\multicolumn{2}{c|}{Metric} & MSE & MAE& MSE  & MAE& MSE  & MAE& MSE  & MAE\\
\midrule
\multirow{4}{*}{\rotatebox{90}{$ETTm2$}}
& 96  &\textbf{0.179} &\textbf{0.245}   & 0.189& 0.296& 0.200 & 0.296& 0.213& 0.324\\
& 192 &\textbf{0.233} &\textbf{0.309}   & 0.248& 0.318& 0.257 & 0.323& 0.258& 0.349\\
& 336  &\textbf{0.303} &\textbf{0.356}  & 0.355& 0.404& 0.334 & 0.376& 0.344& 0.400\\
& 720 &\textbf{0.387} &\textbf{0.413}   & 0.445& 0.458& 0.463 & 0.485&  0.479& 0.506\\
\midrule
\multirow{4}{*}{\rotatebox{90}{$Exchange$}}
& 96  &\textbf{0.091} &\textbf{0.224}   & 0.104& 0.225& 0.114& 0.246& 0.108& 0.236 \\
& 192 &\textbf{0.219} &\textbf{0.346}   & 0.237& 0.380& 0.242& 0.363& 0.244& 0.375 \\
& 336 &\textbf{0.412} &\textbf{0.502}   & 0.446& 0.493& 0.438& 0.485& 0.479& 0.506 \\
& 720 &\textbf{0.690} &\textbf{0.647}   & 1.425& 0.983& 1.509& 0.907& 1.041& 0.868 \\
\bottomrule
\end{tabular}
}
\label{tab:ablation}
\vskip -0.1in
\end{table}

 Table \ref{tab:ablation} shows the results of the ablation experiments. We observe that each component in the model produces a positive effect on the performance of TreeDRNet. For dataset $ETTm2$, tree structure achieves the largest improvement compared to the other two components. In data set $Exchange$, the improvement of the accuracy of the three components is similar. It is worth noting that for long-range 720 predictions in $Exchange$, any two combination of components does not achieve excellent results, showing that the prediction results are better when the three components work together.

 \section{Ablation Study: Double Residual Link Structure}
\begin{figure}[t]
  \centering
  \includegraphics[width=1\linewidth]{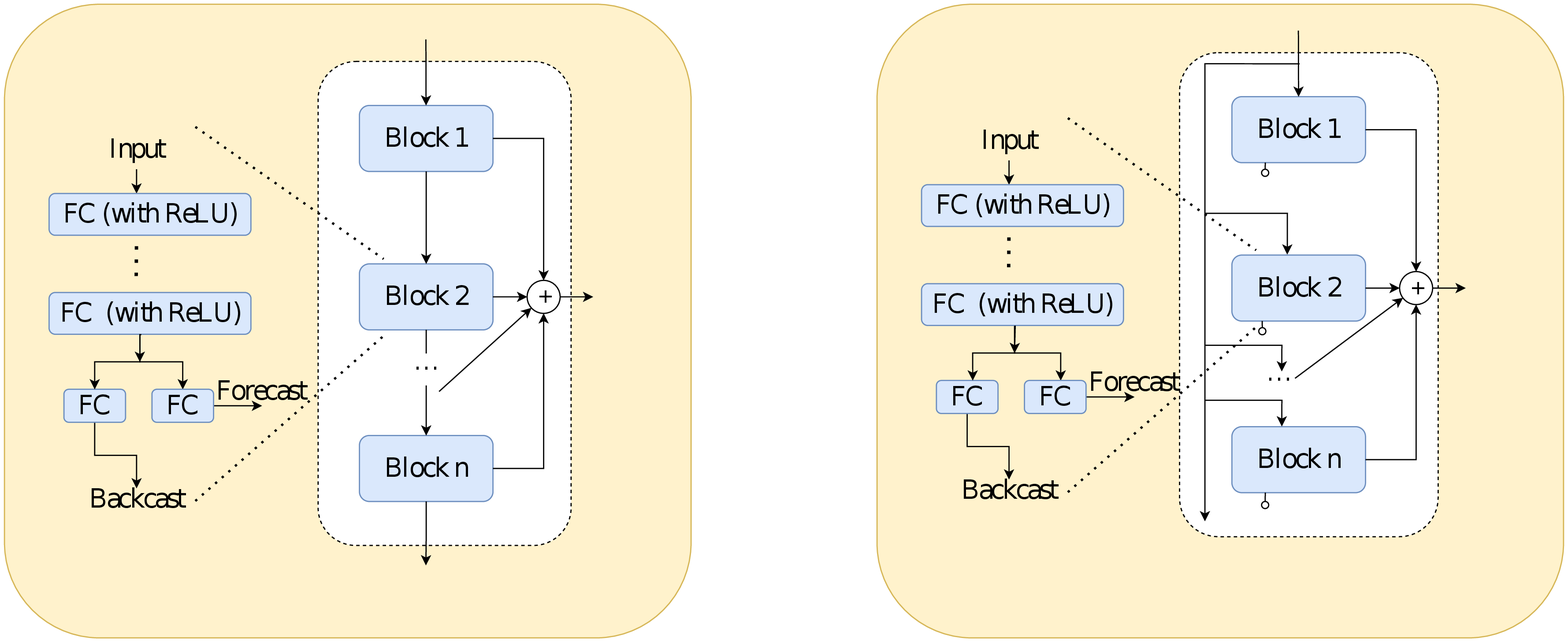}
  \caption{ (a) NoResidual block~~~~~~~~~~~~~~~~~~~~~~~~~~~~~~~~~~~~~~~~~~ (b) Parallel block~~~~~~~~~~~~}
  \label{fig:ablation-double-residual}
\end{figure}

In this part, we conduct an ablation experiment on two different alternative versions of the double residual link structure in the TreeDRNet. One is the NoResidual version: we delete the residual path in the original structure; the other is the Parallel version: we use the same input $x$ for all blocks in doubly residual structure without the backcast processing. The structure graphs are plotted in Fig.~\ref{fig:ablation-double-residual}. Their forecasting performance comparisons on Exchange and ETTm2 datasets are summarized in Table \ref{tab:ablation_double}. It can be seen that the original double residual linked structure adopted in the proposed TreeDRNet model achieves the best and most stable performance compared to its two alternative versions as shown in  

\begin{table}[h]
\centering
\caption{Forecasting comparisons of TreeDRNet with double residual link structure and the two ablation versions (NoResidual and Parallel) on Exchange and ETTm2 datasets. The best results are highlighted in bold.}
\scalebox{0.85}{
\begin{tabular}{c|c|ccccccccc}
\toprule
\multicolumn{2}{c|}{Model}&\multicolumn{2}{c|}{TreeDRNet}&\multicolumn{2}{c|}{NoResidual}&\multicolumn{2}{c}{Parallel}\\
\midrule
\multicolumn{2}{c|}{Metric} & MSE & MAE& MSE  & MAE& MSE  & MAE\\
\midrule
\multirow{4}{*}{\rotatebox{90}{$ETTm2$}}
& 96  &\textbf{0.179} &\textbf{0.245}   & 0.193& 0.265& 0.181 & 0.258\\
& 192 &\textbf{0.233} &\textbf{0.309}   & 0.263& 0.323& 0.241 & 0.308\\
& 336  &0.303 &0.356  & \textbf{0.302}& \textbf{0.346}& 0.314 & 0.350\\
& 720 &\textbf{0.387} &\textbf{0.413}   & 0.416& 0.434& 0.400 & 0.413\\
\midrule
\multirow{4}{*}{\rotatebox{90}{$Exchange$}}
& 96  &\textbf{0.091} &\textbf{0.224}   & 0.130& 0.246& 0.126& 0.252\\
& 192 &\textbf{0.219} &\textbf{0.346}   & 0.260& 0.365& 0.244& 0.359\\
& 336 &0.412&0.502   & \textbf{0.335}& \textbf{0.429}& 0.552& 0.547\\
& 720 &\textbf{0.690} &\textbf{0.647}   & 0.987& 0.770& 1.210& 0.861\\
\bottomrule
\end{tabular}
}
\label{tab:ablation_double}
\vskip -0.1in
\end{table}

\section{Sensitivity Analysis}\label{app:Sensitivity}

In this part, we examine the performance sensitivity of the hidden layer dimension (shown in Table \ref{tab:different-hidden-units}) and the depth of tree (shown in Table \ref{tab:different-loop}) in the TreeDRNet model.
 
 As shown in table \ref{tab:different-hidden-units}, firstly, with the number of hidden dimension increasing, the prediction accuracy improves. After reaching a certain number (such as 128), the subsequent increase of hidden dimension would make model begin to overfit the training data. Therefore, the tradeoff between the complexity and performance of our model can be controlled to a suitable hidden size.
 The depth of our tree-based model has a similar impact as shown in table \ref{tab:different-loop}. The performance reaches a saturation status when the depth increases to some point, which shows that our model does not require a large depth to achieve more accurate predictions.
 
\begin{table*}[h]
\centering
\caption{Sensitivity experients of hidden dimension. The best results are highlighted in bold.}\vspace{-1mm}
\scalebox{0.9}{
\begin{tabular}{c|c|cccccccccccccc}
\toprule
\multicolumn{2}{c|}{Hidden Dim}&\multicolumn{2}{c|}{32}&\multicolumn{2}{c|}{64}&\multicolumn{2}{c|}{128}&\multicolumn{2}{c|}{256}&\multicolumn{2}{c}{512}\\
\midrule
\multicolumn{2}{c|}{Metric} & MSE & MAE& MSE  & MAE& MSE  & MAE& MSE  & MAE& MSE  & MAE\\
\midrule
\multirow{4}{*}{\rotatebox{90}{$ETTm2$}}
& 96  &\textbf{0.174} &0.264 & 0.192& 0.292& 0.179 & \textbf{0.245}& 0.175& 0.266& 0.199& 0.291\\
& 192 &0.248 &0.317 & 0.282& 0.352& \textbf{0.233} & \textbf{0.309}& 0.237& 0.309& 0.241& 0.309\\
& 336  &0.308 &0.364 & 0.319& 0.377& \textbf{0.303} & \textbf{0.356}& 0.306& 0.364& 0.286& 0.341\\
& 720 &0.600 &0.557 & 0.547& 0.510& 0.426 &0.436&  \textbf{0.406}& \textbf{0.432}& 0.432& 0.450\\
\midrule
\multirow{4}{*}{\rotatebox{90}{$Exchange$}}
& 96  &0.098 &0.226 & 0.108& 0.235& 0.091& 0.224& \textbf{0.087}& \textbf{0.212} &0.089 &0.216\\
& 192 &0.248 &0.371 & 0.244& 0.367& \textbf{0.219}& \textbf{0.346}& 0.246& 0.380 &0.227 &0.350\\
& 336 &0.448 &0.514 & \textbf{0.348}& \textbf{0.441}& 0.412& 0.502& 0.408& 0.484 &0.447 &0.510\\
& 720 &1.456 &0.862 & 1.215& 0.844& \textbf{0.690}& \textbf{0.647}& 1.848& 0.971 &1.857 &0.973\\
\bottomrule
\end{tabular}
}
\label{tab:different-hidden-units}
\vskip -0.1in
\end{table*}
\begin{table*}[h]
\centering
\caption{Sensitivity experiments of different tree depth. The best results are highlighted in bold.}\vspace{-1mm}
\scalebox{0.9}{
\begin{tabular}{c|c|ccccccccc}
\toprule
\multicolumn{2}{c|}{Tree depth}&\multicolumn{2}{c|}{1}&\multicolumn{2}{c|}{2}&\multicolumn{2}{c}{3}\\
\midrule
\multicolumn{2}{c|}{Metric} & MSE & MAE& MSE  & MAE& MSE  & MAE\\
\midrule
\multirow{4}{*}{\rotatebox{90}{$ETTm2$}}
& 96  &0.203 &0.283 & \textbf{0.179}& \textbf{0.245}& 0.189 & 0.285\\
& 192 &0.246 &0.317 & \textbf{0.233}& \textbf{0.309}& 0.279 & 0.347\\
& 336 &0.320 &0.374 & \textbf{0.303}& \textbf{0.356}& 0.314 & 0.364\\
& 720 &0.445 &0.473 & 0.426& 0.436& \textbf{0.393} &\textbf{0.424}\\
\midrule
\multirow{4}{*}{\rotatebox{90}{$Exchange$}}
& 96  &0.136 &0.270 & 0.091& 0.224& \textbf{0.089}& \textbf{0.218}\\
& 192 &0.224 &0.349 & 0.219& 0.346& \textbf{0.198}& \textbf{0.327}\\
& 336 &0.458 &0.524 & \textbf{0.412}& \textbf{0.502}& 0.506& 0.531\\
& 720 &0.768 &0.721 & \textbf{0.690}& \textbf{0.647}& 1.673& 0.916\\
\bottomrule
\end{tabular}
}
\label{tab:different-loop}
\vskip -0.1in
\end{table*}

\section{Comparison of Memory Usage} \label{app:mem}
\begin{figure}[h]
  \centering
  \includegraphics[width=0.4\linewidth]{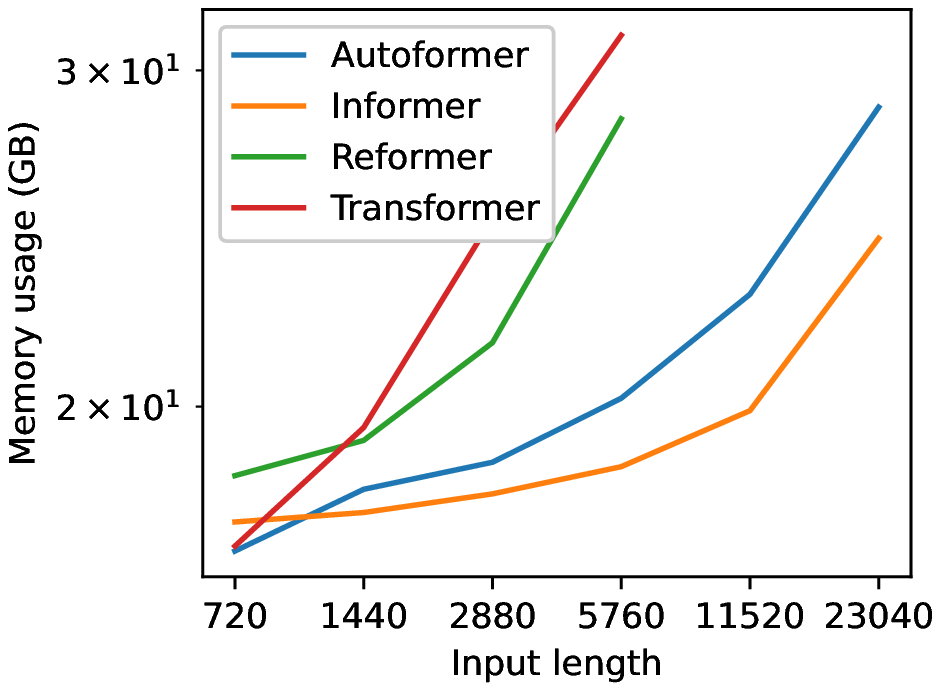} 
  \includegraphics[width=0.4\linewidth]{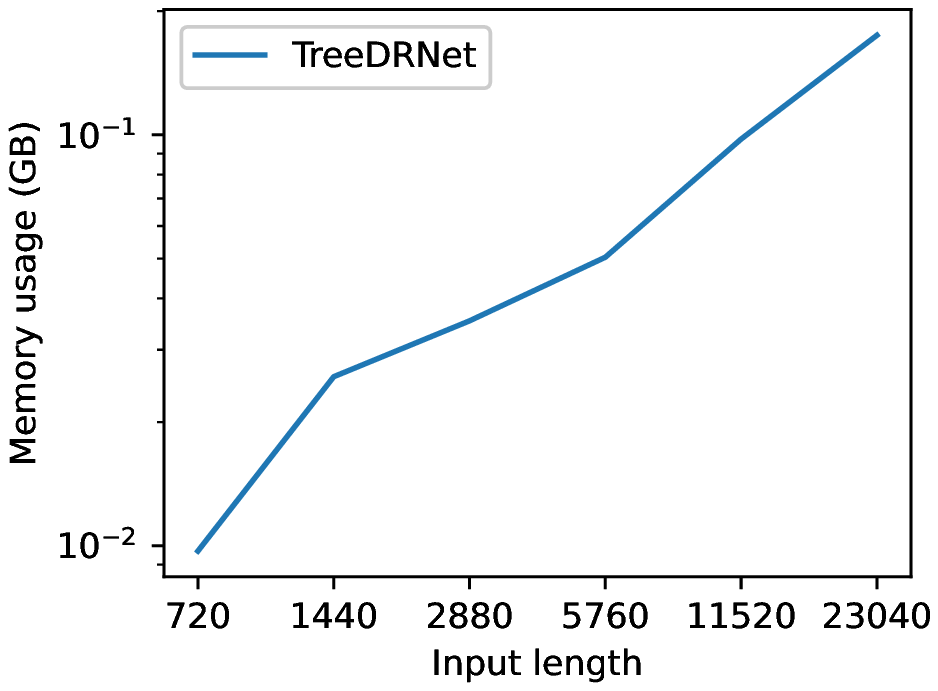}
  \caption{Left: the memory usage of Autoformer, Informer, Reformer and Transformer models with prolonging input length. Right: the memory usage of TreeDRNet with prolonging input length.}
  \label{fig_memory}
\end{figure}

In this part, we examine the memory usage of different SOTA Transformer based models and our TreeDRNet as shown in Fig.~\ref{fig_memory}. It can be seen that the memory usage of our TreeDRNet model is two orders of magnitude smaller than various transformer counterparts. Furthermore, the memory usage of our TreeDRNet increases \textbf{linearly} with prolonging input length, compared to $O(nlogn)$ increasing rate of the Transformer based models. 
 
\newpage
\section{Robustness Theoretical Analysis}
Since we have thoroughly discussed the robustness of TreeDRNet in Section Motivation and Theoretical Foundation, our focus here is switched to attention layer since it holds state-of-the-art performance for long-term time forecasting tasks. 
Since the self-attention module has achieved great success in CV, NLP, and time series fields, it seems counter-intuitive that it is not robust to noisy input. We show that low hidden dimension and small layer number make the self-attention layer much less robust in most time-series transformer implementations than its NLP counterpart \cite{hsiehetal2019robustness}. 

\paragraph{Sensitivity of Self-Attention Layers}
Let us consider the simplest case of one self-attention layer with one single head. Assume our input length is $n$ and the hidden dimension is $d$, denoted by $x_{1}, \ldots, x_{n} \in R^{d}$. We use $W^{Q}$, $W^{K}$, and $W^{V} \in R^{d \times k}$ to denote the query, key and value transformations. The contribution of each $j$ to $i$ can be computed as 
$s_{i j}^{\prime}=x_{i}^{T} W^{Q}\left(W^{K}\right)^{T} x_{j}$. The $i$-th position of the output for the next layer can be calculated as 
$X_{i}=\sum_{j} \frac{e^{s_{i j}}}{\sum_{k} e^{s_{i k}}}\left(W^{V} x_{j}\right)$.
Here, we assume a litter noise is added to a particular index $k$, such that $x_k$ is changed to $x_k+\delta x$ while all others remain unchanged. We then study how much this noise injection will affect the output. For a particular $i\neq j$, the $s_{ij}$ is changed by one term as the following formula
$$s_{i j}^{\prime}=\left\{\begin{array}{ll}s_{i j} & \text { if } j \neq k\\ s_{i j}+x_{i}^{T} W^{Q}\left(W^{K}\right)^{T} \Delta x & \text { if } j=k\end{array}\right..$$
Here we can see the changed term is the inner product between $x_i$ and a fixed vector $ W^{Q}\left(W^{K}\right)^{T} \Delta x$. For the application in NLP, the embeddings ${x_{i}}_{j=1}^{n}$ are scattered enough over the space, and the inner product cannot be large for all ${x_{i}}_{j=1}^{n}$. But for application in time series, it usually not the case since the point embedings are much more similar. Moreover, even we assume such sparsity in time series the same as in NLP, we can still show its week robustness.
For instance, we can prove the sparsity under some distributional assumption on ${x_i}$ as
\begin{theorem}\label{eq:thm1} Assume $\|\Delta x\| \leq \delta$ and $\left\{x_{i}\right\}_{i=1}^{n}$ are $d$-dimensional vectors uniformly distributed on the unit sphere, then $\mathbb{E}\left[\left|s_{i \bar{j}}^{\prime}-s_{i j}\right|\right] \leq \frac{C \delta}{\sqrt{d}}$ with $C=$ $\left\|W^{Q}\right\|\left\|W^{K}\right\|$ and $\mathbb{P}\left(\left|s_{i \bar{j}}^{\prime}-s_{i j}\right| \geq \epsilon\right) \leq \frac{C \delta}{\epsilon \sqrt{d}}$.
\end{theorem}
\begin{proof}
The value $\mathbb{E}\left[|s_{i j}^{\prime}-s_{i j}|\right]=\mathbb{E}\left[|x_{i}^{T} z|\right]$, where $z=W^{Q}\left(W^{K}\right)^{T} \Delta x$ is a fixed vector. Then, it is easy to derive $\|z\| \leq\left\|W^{Q}\right\|\left\|W^{K}\right\| \delta$. The remaining task follows by applying the Hoeffiding inequality on $x_{i}^{T} z$.

\end{proof}

In natural language processing (NLP) application, the norm of $W^{Q}$, $W^{K}$ are are typically upper bounded and the hidden dimensional $d$ is relatively large (e.g., 768). According to Theorem \ref{eq:thm1}, a large amount of perturbations are negligible. 
However, for the application in time series, the overfitting issue forces us to choose a small hidden dimensional $d$ (e.g., 32). The robustness deteriorates significantly. 

Moreover, as shown in \cite{hsieh-etal-2019-robustness} for self-attention layers in BERT, the distribution of change on embedding is sparse after going through the first self-attention layer and then gradually propagates to the following layers. The 12 attention layers in BERT can provide extra robustness. 
Nevertheless, for the application in time series, we choose a small number of attention layer (1 or 2) to avoid overfitting, which differs from the situation in NLP. 
 
\section{Robustness Experimental Analysis: Attacks}
In this part, we conduct a thorough noise attack experiment to show the robustness of the proposed TreeDRNet model compared to SOTA transformer based models. We consider three different attacking policies borrowed from the noisy injection schemes of time series anomaly detection field~\cite{Carmona2021NeuralCA}.

\begin{figure}[t]
\centering
\begin{minipage}[t]{0.32\textwidth}
\centering
\includegraphics[width=\linewidth]{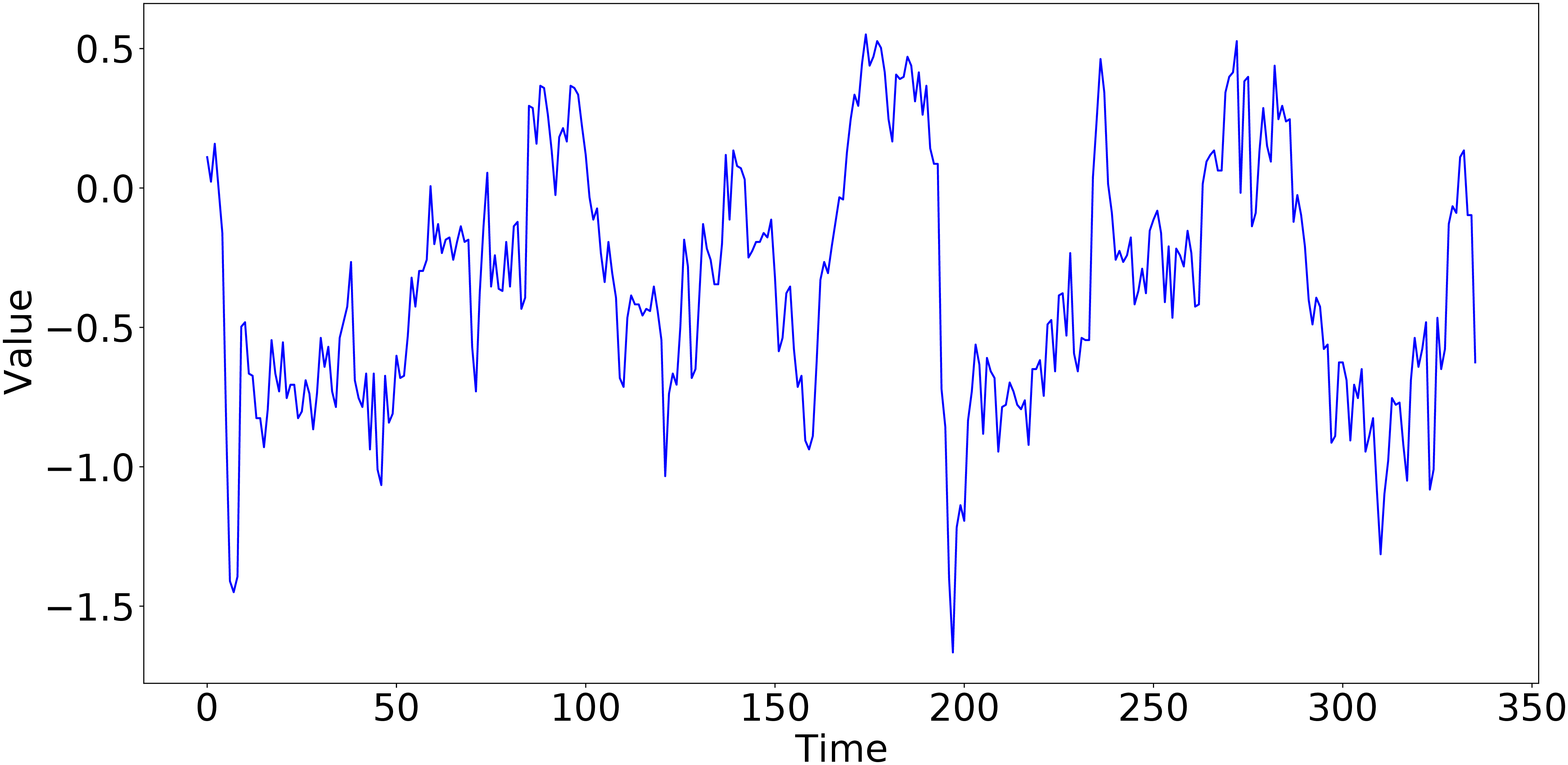}
\end{minipage}
\begin{minipage}[t]{0.32\textwidth}
\centering
\includegraphics[width=\linewidth]{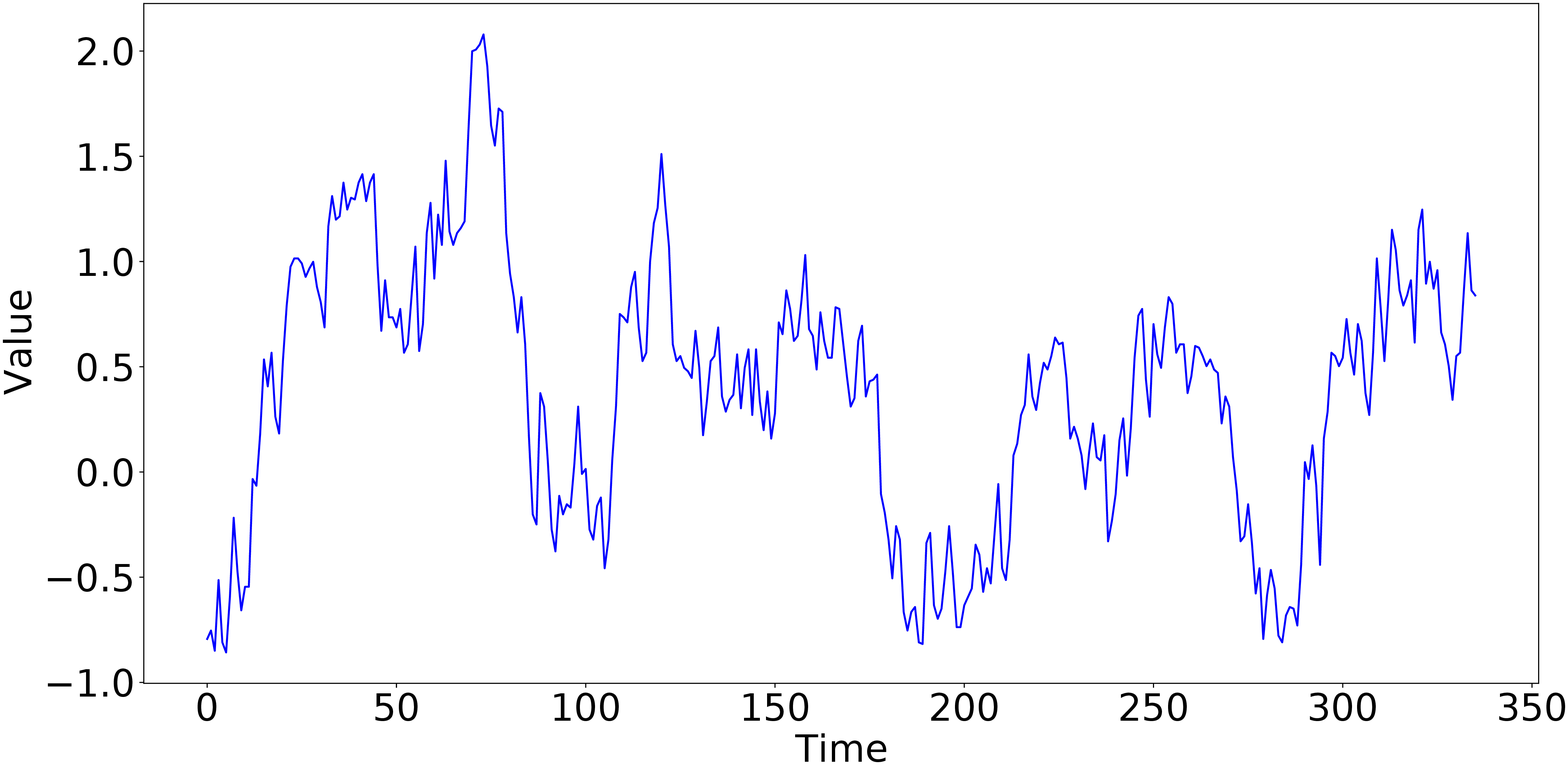}
\end{minipage}
\begin{minipage}[t]{0.32\textwidth}
\centering
\includegraphics[width=\linewidth]{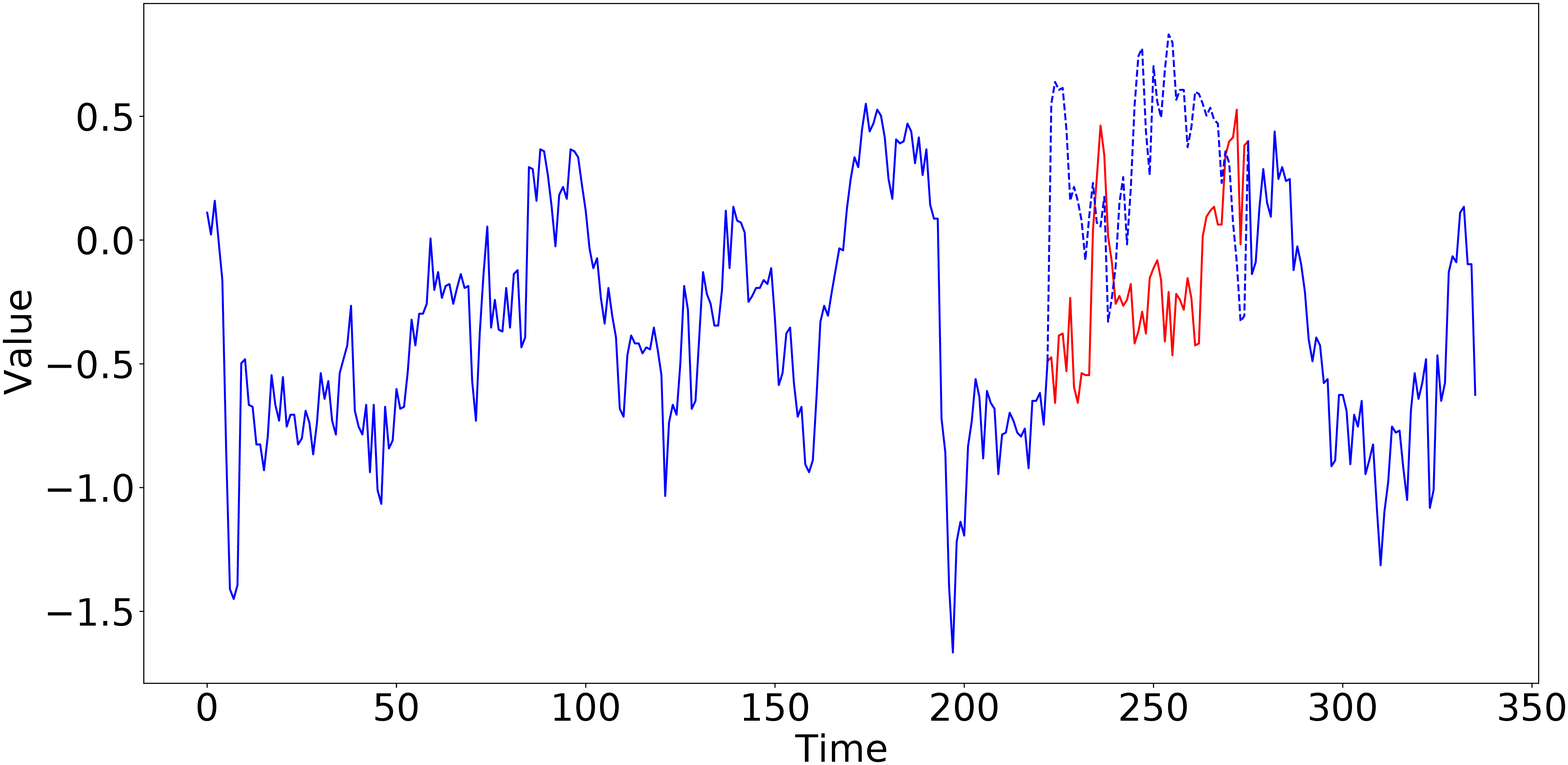}
\end{minipage}
\caption{(a) Series 1~~~~~~~~~~~~~~~~~~~~~~~~~ (b) Series 2~~~~~~~~~~~~~~~~~~~~~~~~~ (c) Series1 COE attack using Series2}
\label{fig:COE}
\end{figure}

\begin{figure}[t]
\begin{minipage}[t]{0.48\textwidth}
\centering
\includegraphics[width=7cm]{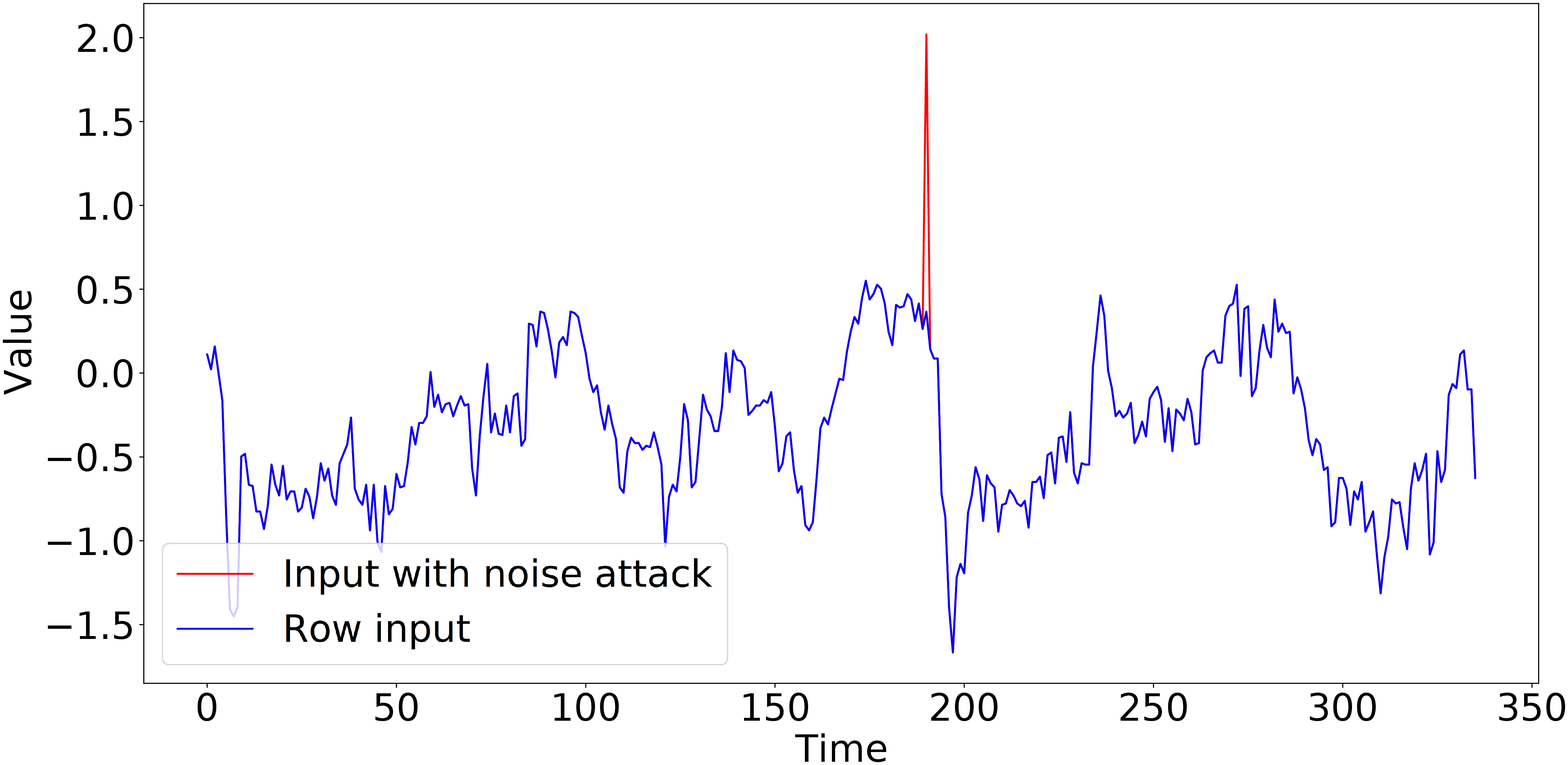}
\end{minipage}
\begin{minipage}[t]{0.48\textwidth}
\centering
\includegraphics[width=7cm]{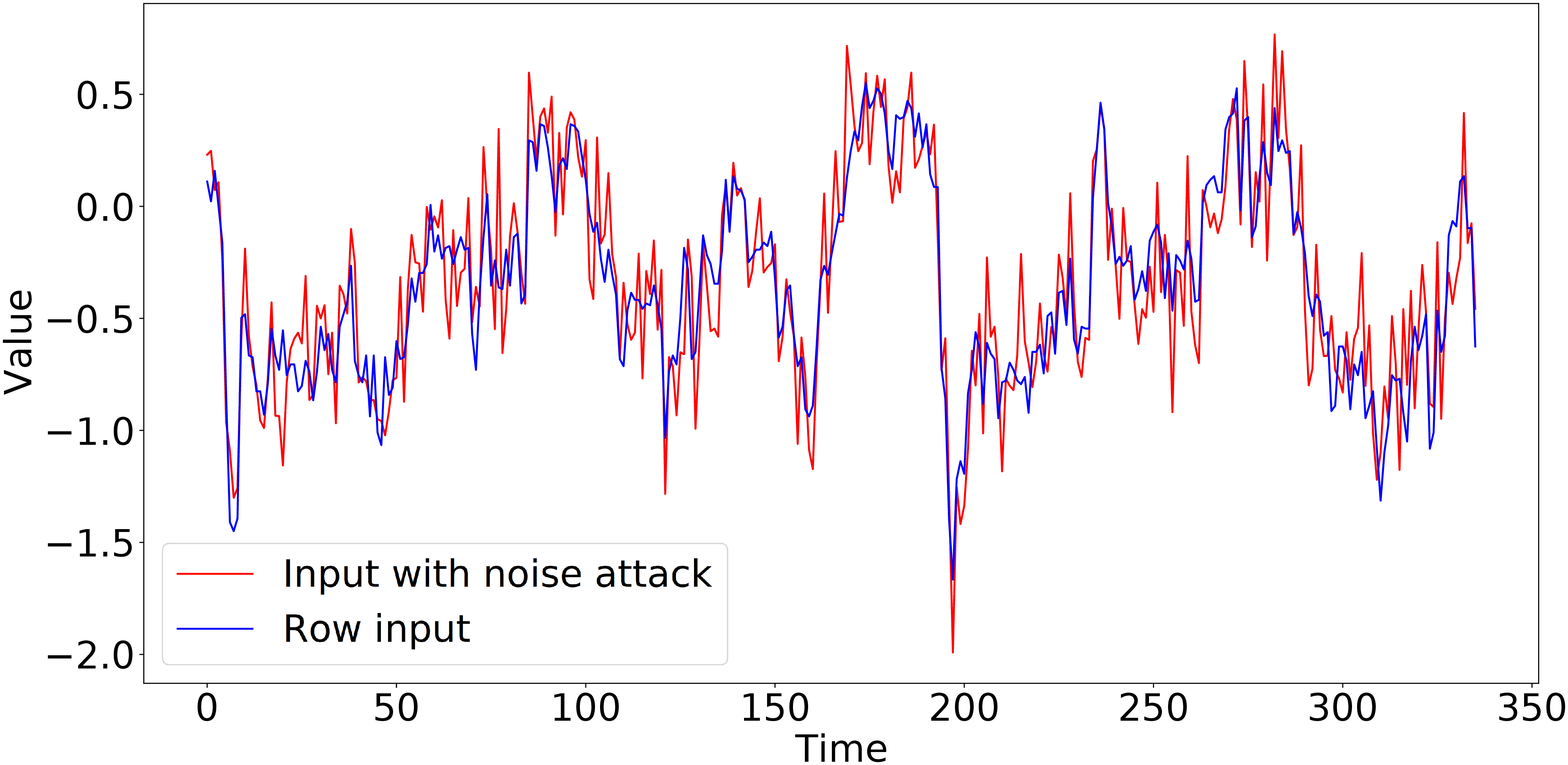}
\end{minipage}
\caption{(a) Series with anomaly injection attack~~~~~~~~ (b) Series with white noise attack}
\label{fig:anomaly}
\end{figure}

\paragraph{Contextual Outlier Exposure Attack (COE)}
Given a data window $\boldsymbol{w}=\left(\boldsymbol{w}^{(c)}, \boldsymbol{w}^{(s)}\right)$, we inject noise into the suspect segment, $\boldsymbol{w}^{(s)}$, by replacing a chunk of its values with values taken from another time series. The replaced values in $\boldsymbol{w}^{(s)}$ will most likely break the temporal relation with their neighboring context, therefore creating an out of distribution noisy input. In our implementation, we apply COE at training time by selecting all examples in a minibatch and permuting a random length in $40\%$ of the total length. Since we conduct multivariate time series forecasting here, we select all of the dimensions in which a window of signal is swapped. A sample sequence after attacking is shown in Figure \ref{fig:COE}

\begin{figure}[t]
\begin{minipage}[t]{0.48\textwidth}
\centering
\includegraphics[width=7cm]{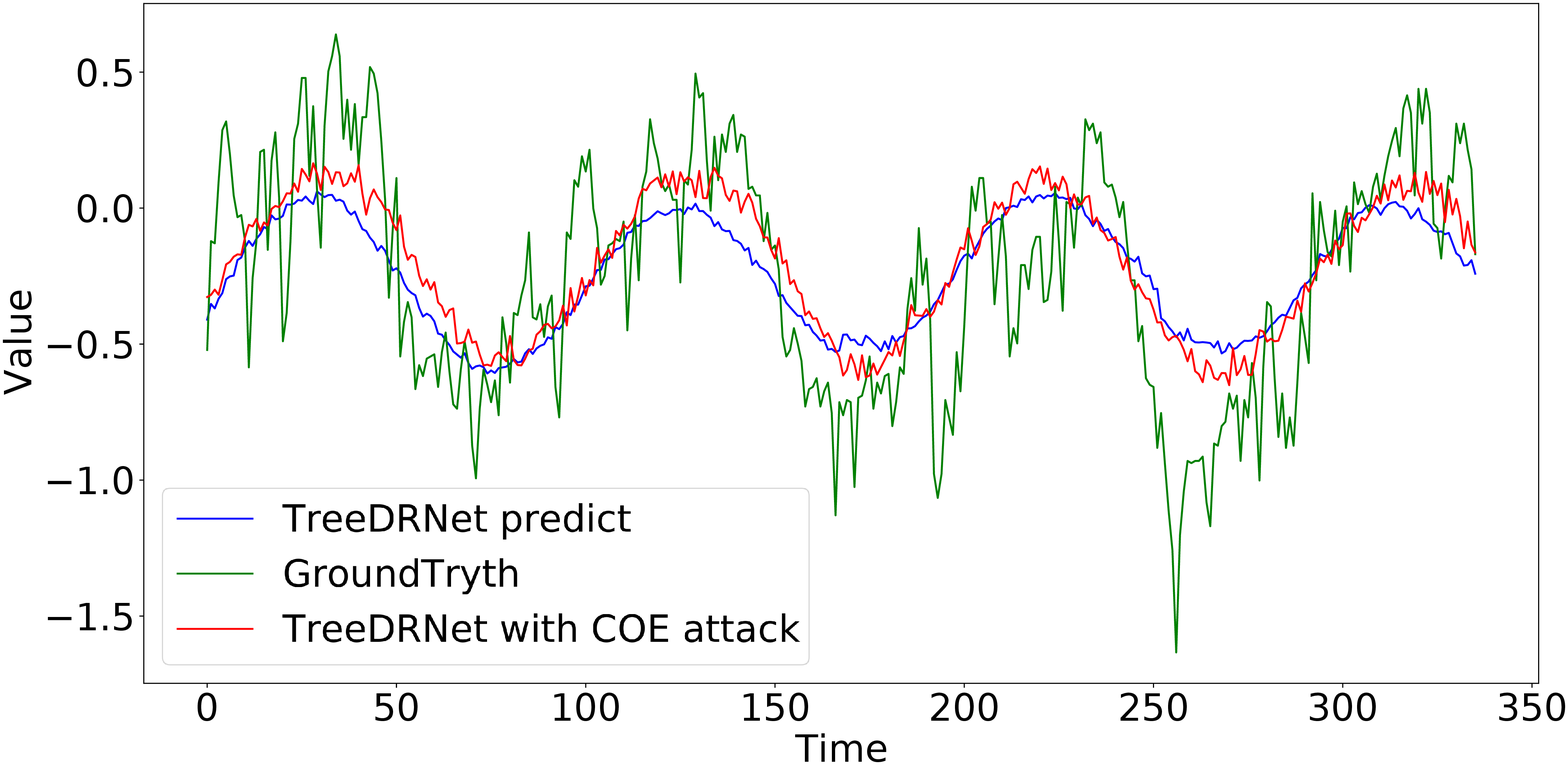}
\end{minipage}
\begin{minipage}[t]{0.48\textwidth}
\centering
\includegraphics[width=7cm]{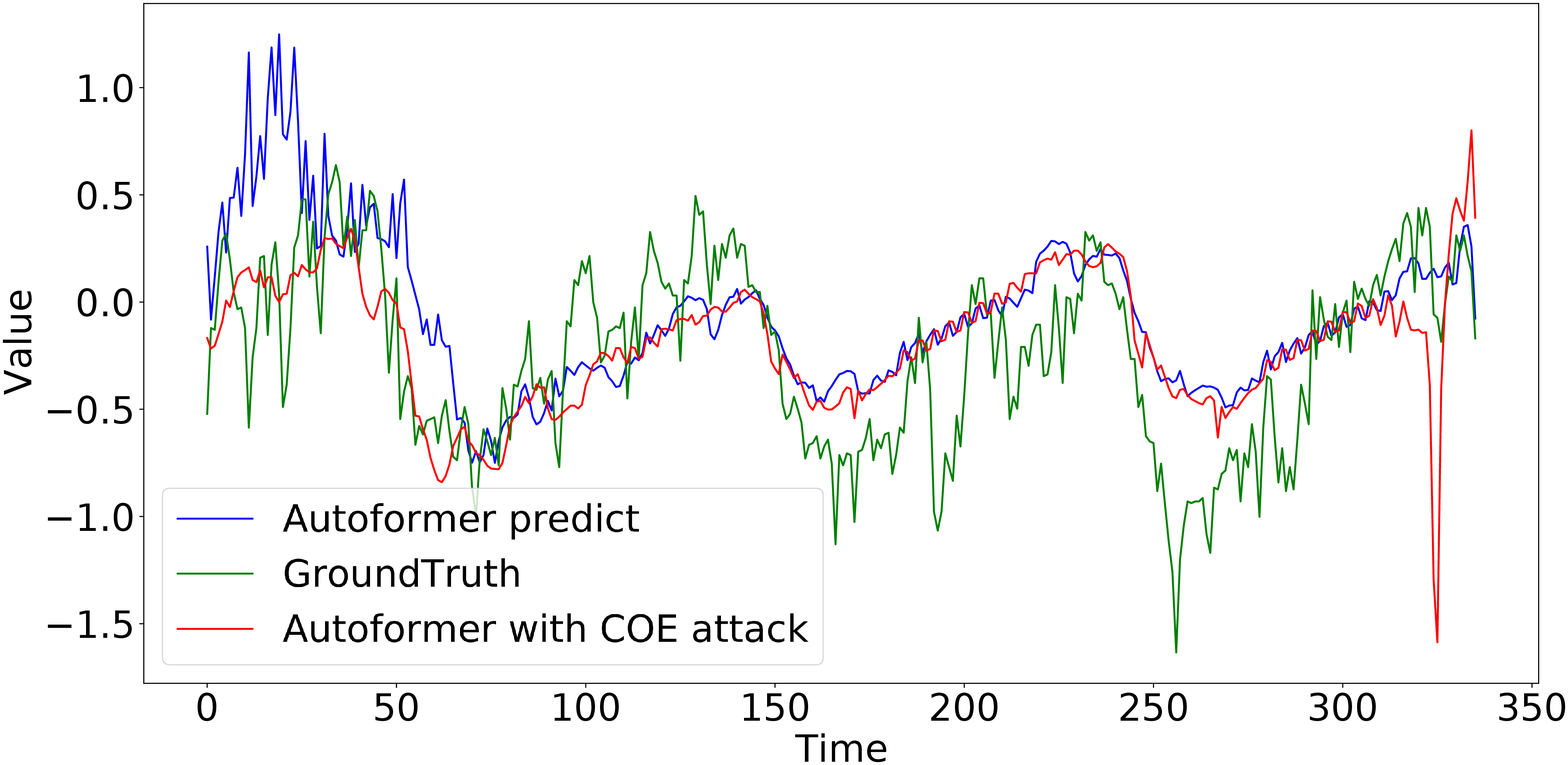}
\end{minipage}
\caption{(a) TreeDRNet with COE attack~~~~~~~~ (b) Autoformer with COE attack}
\label{fig:predict_coe}
\end{figure}

\begin{figure}[htbp]
\begin{minipage}[t]{0.48\textwidth}
\includegraphics[width=7cm]{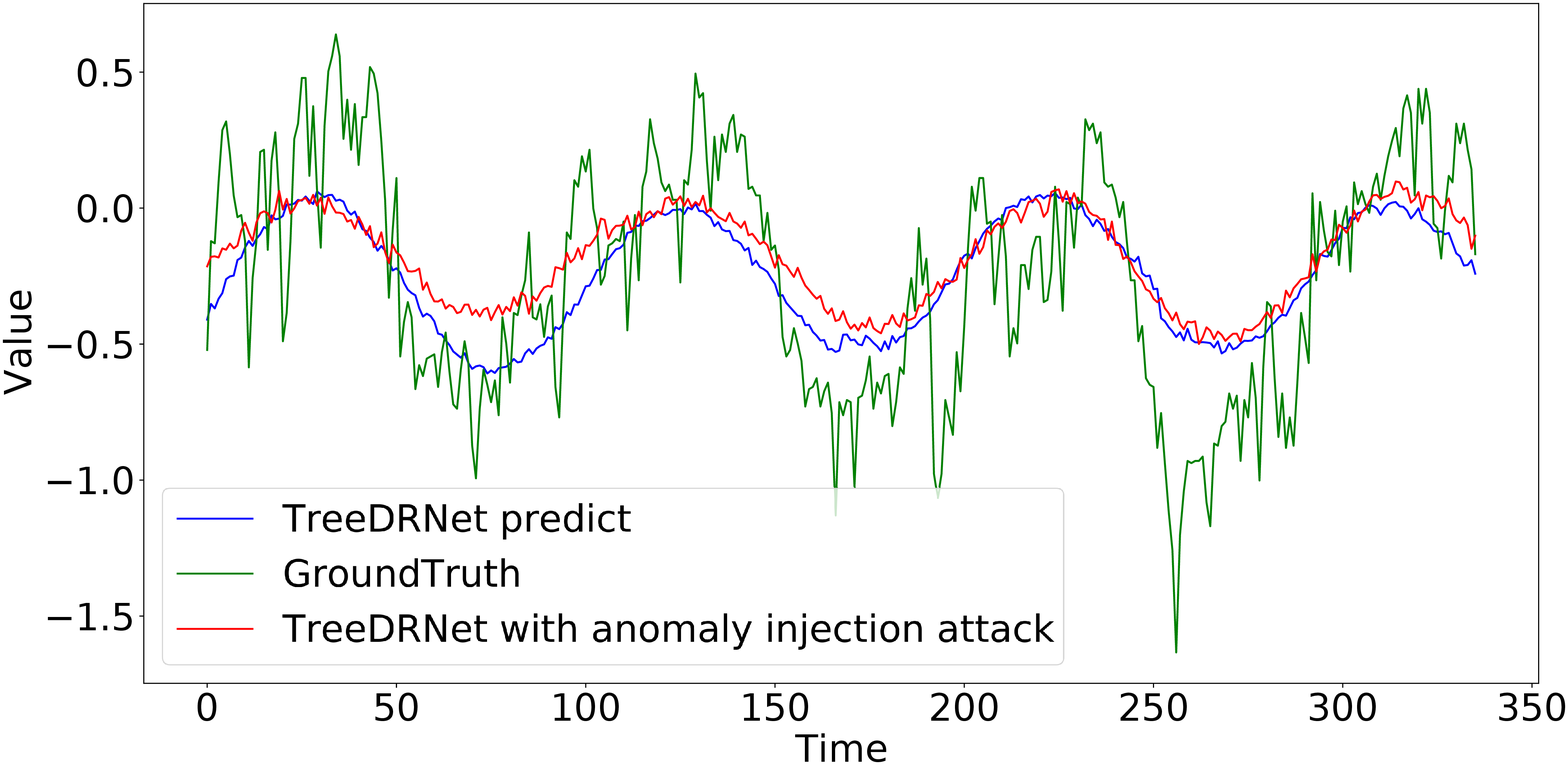}
\end{minipage}
\begin{minipage}[t]{0.48\textwidth}
\includegraphics[width=7cm]{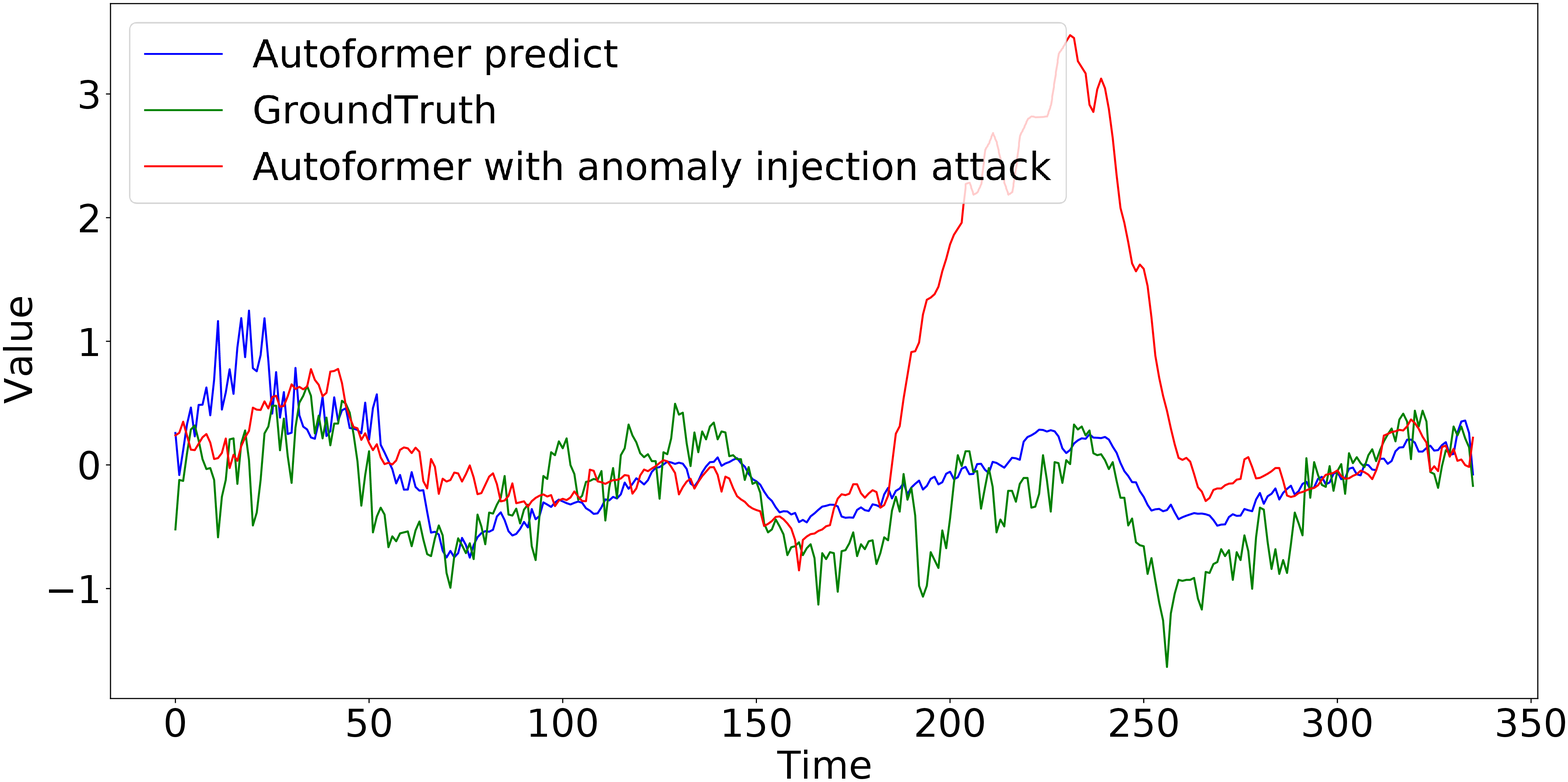}
\end{minipage}
\caption{(a) TreeDRNet with anomaly injection attack (b) Autoformer with anomaly injection attack}
\label{fig:predict_anomaly}
\end{figure}

\begin{figure}[htbp]
\begin{minipage}[t]{0.48\textwidth}
\centering
\includegraphics[width=7cm]{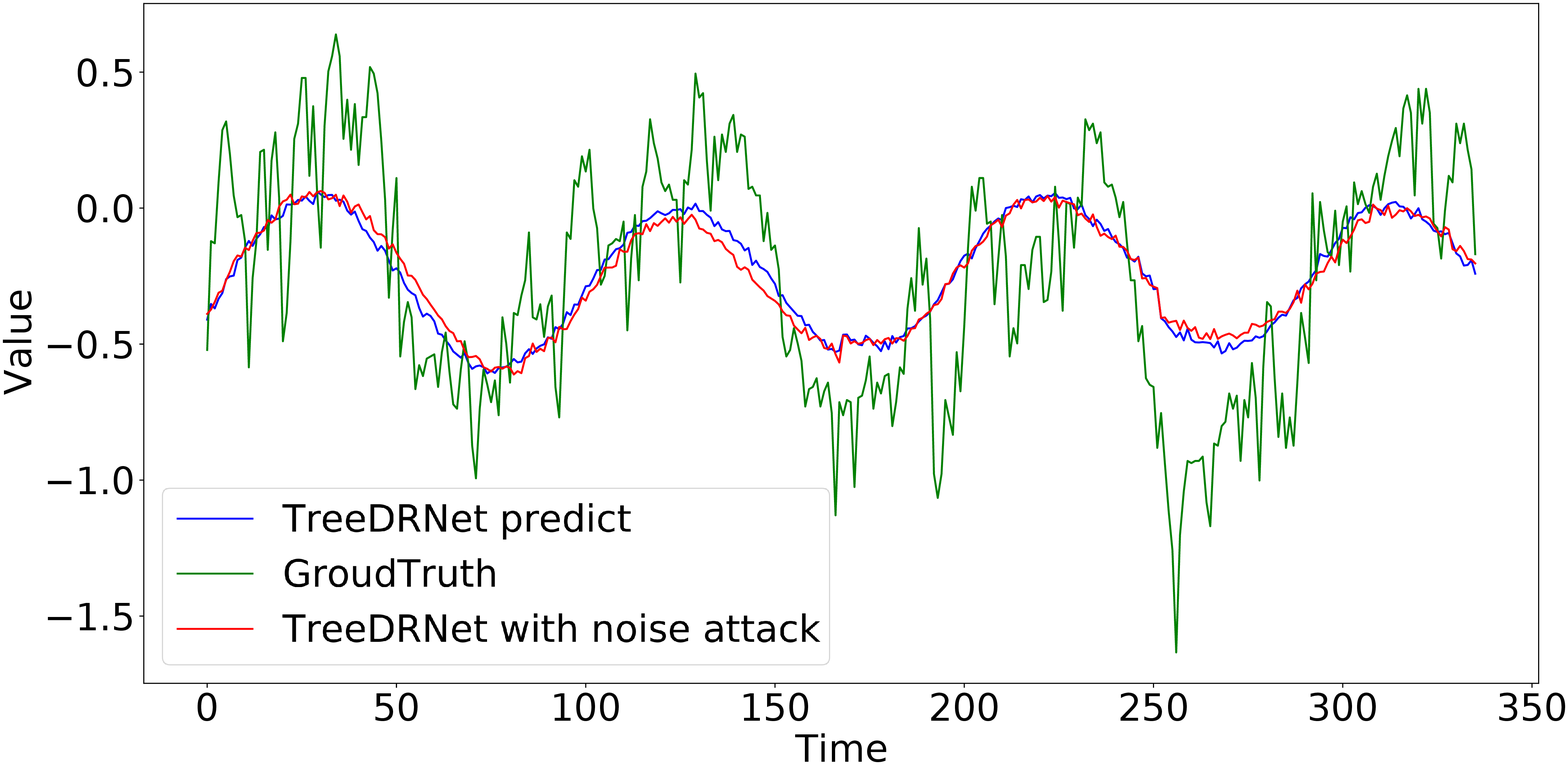}
\end{minipage}
\begin{minipage}[t]{0.48\textwidth}
\centering
\includegraphics[width=7cm]{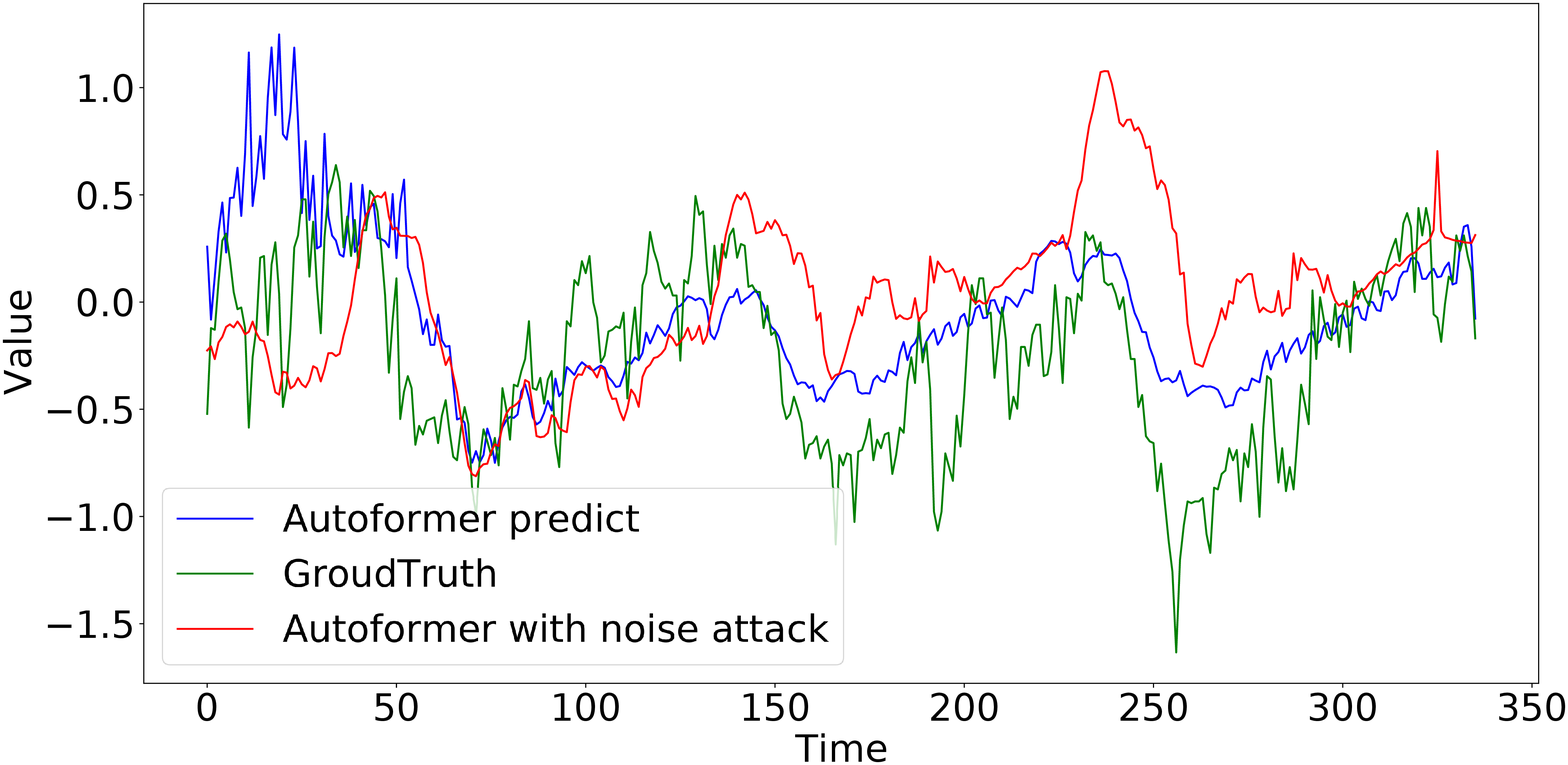}
\end{minipage}
\caption{(a) TreeDRNet with white noise attack~~~~ (b) Autoformer with white noise attack}
\label{fig:predict_white_noise}
\end{figure}

\paragraph{Anomaly Injection Attack}
 We use a simple single Point Outliers injection attack method: at a subset of randomly selected points we add (or subtract) one spike point to each input sequence. The spike is proportional to the inter-quartile range of the points surrounding the spike location, in this experiment, we chose a range of 5-20 times the maximum value of the sequence as spike. Like for $\mathrm{COE}$, in multivariate time series we select all of dimensions on which we add the spike. This attack is used for measure model's robustness with spike noisy attacks. A sample sequence after attacking is shown in Figure \ref{fig:anomaly}(a).

\paragraph{White Noise Attack}
We use a white noise attack method: at a subset of randomly selected points we add random white noise signal in $0.4*N(0,1)$ to the time series. Like for $\mathrm{COE}$, we add the white noise in all of dimensions to test the robustness for white noise attacks. 
A sample sequence after attacking is shown in Figure \ref{fig:anomaly}(b).

We summarize the forecasting performance on the ETT dataset of the proposed TreeDRNet model and the SOTA Autoformer model under the aforementioned three different attack methods in Table~\ref{tab:attack-result}. It can be seen that our TreeDRNet model shows superior robustness compared with the transformer-based Autoformer model. Facing COE attack, Autoformer has $21.6\%$ performance decrease while the TreeDRNet only decreases $5.8\%$. Facing anomaly injection attack, Autoformer has $16.7\%$. performance decrease while the TreeDRNet only decreases $8.1\%$. Facing white noise attack, the Autoformer has $24.9\%$. performance decrease while the TreeDRNet only decreases $5.1\%$. Experimentally, when facing prolonged data input or noisy injection attacks, our TreeDRNet model performs consistently robustly, supporting our robust algorithm claim. For illustrated comparison between TreeDRNet and Autoformer, 
a sample forecasting sequence after COE attacked training is shown in Figure \ref{fig:predict_coe}; a sample forecasting sequence after anomaly attacked training is shown in Figure \ref{fig:predict_anomaly}; and a sample forecasting sequence after white noise attacked training is shown in Figure \ref{fig:predict_white_noise}.

\begin{table}[h]
\renewcommand\arraystretch{1.2}
\centering
\caption{Forecasting comparisons of TreeDRNet and Autoformer under different attack methods on the $ETT$ dataset.}
\scalebox{0.85}{
\begin{tabular}{c|c|ccccccccccc}
\toprule
\multicolumn{2}{c|}{Model}&\multicolumn{2}{c|}{Benchmark}&\multicolumn{2}{c|}{COE}&\multicolumn{2}{c|}{Anomaly injection}&\multicolumn{2}{c}{White noise}\\
\midrule
\multicolumn{2}{c|}{Metric} & MSE & MAE & MSE  & MAE& MSE  & MAE& MSE  & MAE\\
\midrule
\multirow{4}{*}{\rotatebox{90}{TreeDRNet}}
& 96  &\textbf{0.179} &\textbf{0.245}   & 0.197 & 0.276 & 0.223 & 0.320 & 0.202 & 0.301\\
& 192 &\textbf{0.233} &\textbf{0.309}   & 0.253 & 0.313 & 0.258 & 0.327 & 0.249 & 0.318\\
& 336  &\textbf{0.303} &\textbf{0.356}  & 0.310 & 0.379 & 0.296 & 0.348 & 0.311 & 0.365\\
& 720 &\textbf{0.387} &\textbf{0.413}   & 0.395 & 0.427 & 0.388 & 0.402 & 0.389 & 0.406\\
\midrule
\multirow{4}{*}{\rotatebox{90}{Autoformer}}
& 96  &\textbf{0.255} &\textbf{0.339}   & 0.303 & 0.376 & 0.295 & 0.316 & 0.319 & 0.333\\
& 192 &\textbf{0.281} &\textbf{0.340}   & 0.348 & 0.425 & 0.353 & 0.371 & 0.386 & 0.421\\
& 336 &\textbf{0.339} &\textbf{0.372}   & 0.393 & 0.468 & 0.400 & 0.459 & 0.426 & 0.476\\
& 720 &\textbf{0.422} &\textbf{0.419}   & 0.526 & 0.536 & 0.502 & 0.544 & 0.536 & 0.563\\
\bottomrule
\end{tabular}
}
\label{tab:attack-result}
\vskip -0.1in
\end{table}


  

\end{document}